\documentclass[10pt]{article}
\usepackage[utf8]{inputenc}
\usepackage[margin=1in]{geometry}

\usepackage{hyperref}       
\usepackage{url}            
\usepackage{booktabs}       
\usepackage{amsfonts}       
\usepackage{nicefrac}       
\usepackage{microtype}      
\usepackage{graphicx}
\usepackage{subcaption}
\usepackage{booktabs} 
\usepackage{amsmath,amsfonts,amssymb,amsthm}
\usepackage{amsfonts}
\usepackage{dsfont}
\usepackage{xcolor}
\usepackage{mathrsfs}
\usepackage{centernot}
\usepackage{mathtools}
\usepackage{mathrsfs}
\usepackage{wrapfig}
\usepackage{authblk}

\usepackage{enumerate}
\usepackage[round]{natbib}   

\usepackage{array, makecell}
\usepackage{multirow}

\usepackage{arydshln}
\usepackage[flushleft]{threeparttable}

\usepackage{soul}
\usepackage{tikz}
\usetikzlibrary{calc}
\allowdisplaybreaks

\usepackage{soul}
\usepackage{arydshln}
\usepackage[flushleft]{threeparttable}

\makeatletter
\newif\if@anonymize

\@anonymizetrue    

\if@anonymize
  \newcommand{\highlight@DoHighlight}{
    \fill [outer sep = -15pt, inner sep = 0pt, color=gray]
          ($(begin highlight)+(0,8pt)$) rectangle ($(end highlight)+(0,-3pt)$) ;
  }

  \newcommand{\highlight@BeginHighlight}{
    \coordinate (begin highlight) at (0,0) ;
  }

  \newcommand{\highlight@EndHighlight}{
    \coordinate (end highlight) at (0,0) ;
  }

  \newdimen\highlight@previous
  \newdimen\highlight@current
  \newlength{\item@width}

  \DeclareRobustCommand*\anonymize{%
    \SOUL@setup
    \def\SOUL@preamble{%
      \begin{tikzpicture}[overlay, remember picture]
        \highlight@BeginHighlight
        \highlight@EndHighlight
      \end{tikzpicture}%
    }%
    \def\SOUL@postamble{%
      \begin{tikzpicture}[overlay, remember picture]
        \highlight@EndHighlight
        \highlight@DoHighlight
      \end{tikzpicture}%
    }%
    \def\SOUL@everyhyphen{%
      \discretionary{%
        \SOUL@setkern\SOUL@hyphkern
        \SOUL@sethyphenchar
        \tikz[overlay, remember picture] \highlight@EndHighlight ;%
      }{%
      }{%
        \SOUL@setkern\SOUL@charkern
      }%
    }%
    \def\SOUL@everyexhyphen##1{%
      \SOUL@setkern\SOUL@hyphkern
      \settowidth{\item@width}{##1}%
      \makebox[\item@width]{}%
      \discretionary{%
        \tikz[overlay, remember picture] \highlight@EndHighlight ;%
      }{%
      }{%
        \SOUL@setkern\SOUL@charkern
      }%
    }%
    \def\SOUL@everysyllable{%
      \begin{tikzpicture}[overlay, remember picture]
        \path let \p0 = (begin highlight), \p1 = (0,0) in \pgfextra
          \global\highlight@previous=\y0
          \global\highlight@current =\y1
        \endpgfextra (0,0) ;
        \ifdim\highlight@current < \highlight@previous
          \highlight@DoHighlight
          \highlight@BeginHighlight
        \fi
      \end{tikzpicture}%
      \settowidth{\item@width}{\the\SOUL@syllable}%
      \makebox[\item@width]{}%
      \tikz[overlay, remember picture] \highlight@EndHighlight ;%
    }%
    \SOUL@
  }
\else
  \newcommand{\anonymize}[1]{#1}
\fi
\makeatother

\title{Spatial-Temporal-Textual Point Processes for\\Crime Linkage Detection}
\author[1]{Shixiang Zhu}
\author[1]{Yao Xie}
\affil[1]{School of Industrial and Systems Engineering, Georgia Institute of Technology}

\date{}  

\begin{document}

\maketitle

\begin{abstract}
Crimes emerge out of complex interactions of human behaviors and situations. Linkages between crime incidents are highly complex. Detecting crime linkage given a set of incidents is a highly challenging task since we only have limited information, including text descriptions, incident times, and locations. In practice, there are very few labels. We propose a new statistical modeling framework for \textit{spatio-temporal-textual} data and demonstrate its usage on crime linkage detection. We capture linkages of crime incidents via multivariate marked spatio-temporal Hawkes processes and treat embedding vectors of the free-text as {\it marks} of the incident, inspired by the notion of \textit{modus operandi} (M.O.) in crime analysis. Numerical results using real data demonstrate the good performance of our method as well as reveals interesting patterns in the crime data: the joint modeling of space, time, and text information  enhances crime linkage detection compared with the state-of-the-art, and the learned spatial dependence from data can be useful for police operations.
\end{abstract}

\section{Introduction}
\label{sec:introduction}

Spatio-temporal-textual incident data are ubiquitous in modern applications, such as social media posts, electronic health records, and crime incidents. Such incidents data typically include time, location of the incidents, and marks, which include categorical or more detailed descriptions of the incidents. One essential task in analyzing such data is discovering patterns from massive incident data and identifying related incidents. Here, we focus on a particular application arising from police data analysis to identify \textit{crime linkage} from police reports. 

Crime linkage detection plays a vital role in police investigations, aiming to identify a series of incidents committed by a single perpetrator or the same criminal group. The result can help police narrow down the field of search and allocate the workforce more efficiently. Crime linkage detection is usually done by finding a similar \textit{modus operandi} (\emph{M.O.}), typically, using physical or other credible evidence, which are observable traces of the perpetrator such as clothes, fingerprints, DNA, ways to enter the houses and tools, as well as witness statements [\cite{WoBuHo2007,BoJo2016}]. This process is not automatic and usually laborious and requires particular domain knowledge and crime analysis. 

\begin{figure}[!t]
\centering
\includegraphics[width=.5\linewidth]{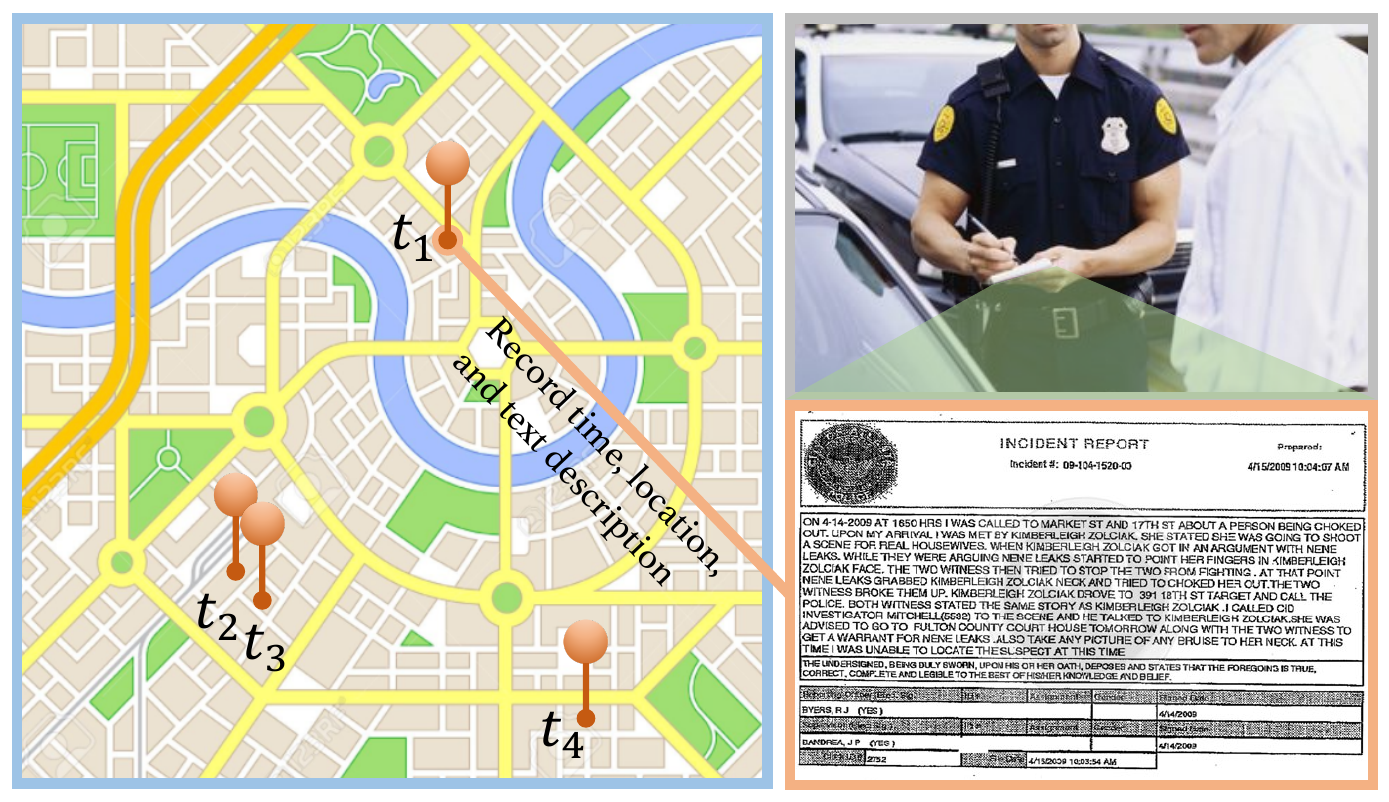}
\caption{An illustrative example of a crime series consists of four crime incidents. Each crime report includes the occurrence time, location, and descriptive text that may contain important information about the incident.
}
\label{fig:intro-exp}
\end{figure}

There is an opportunity to detect crime linkage using data: a wealth of police report data containing extensive information about crime incidents. An illustrative example is shown in Figure~\ref{fig:intro-exp}, which shows a series of crime incidents.
Such a 911 call-for-service report records information about police incidents: when a 911 call is initiated, and a unique incident ID is created. A police officer is dispatched to the scene to investigate the incident and electronically enter the incident's information into the report, which contains structured and unstructured data. The structured data include time, location (street and actual longitude and latitude), and crime category. The unstructured data include narratives and free-text that records interviews with the witnesses or descriptions of the scene. Thus, the police report data is a type of \textit{spatio-temporal} incident data with {\it marks}.

In this paper, we present a framework for modeling crime incidents data, referred to as the \textit{spatio-temporal-textual point process} (\texttt{STTPP}) model, based on which we can detect crime linkages without fully labeled data. Hawkes processes [\cite{Hawkes1971}] have been used extensively to study various topics, including crime [\cite{mohler2011self}], social media [\cite{lai2016topic}], and earthquake prediction [\cite{fox2016spatially}]. In our model, each basic police patrolling geographical unit (\emph{beat}) is regarded as a node in a network associated with a marked Hawkes process. We jointly model spatio-temporal and textual information by incorporating the text as marks of the incidents. To achieve this, we first extract the information from the free-text using the \textit{Bag-of-Words} representation and then map the representation into embedding vectors, which can be viewed as extracted \emph{M.O.} of the incident from the free-text. The embedding is performed by the regularized Restricted Boltzmann Machine (RBM) with {\it keywords selection}. RBM is commonly used as a generative artificial neural network, which we adopt here to capture the joint distribution of keywords and embedding vectors. We further design a new regularization function to perform keyword selection in RBM by penalizing the total probability of the keywords being selected in the model. The keyword selection plays an important role in crime linkage detection because crime series are typically linked via a small set of keywords in the documents. Without performing keyword selection, the model can overfit the training data. Using carefully designed numerical experiments with real-data, we show that our method is highly effective in detecting crime linkages compared with other methods. 

The rest of this paper is organized as follows. We start with a motivating example using real-data for crime linkage detection. Section~\ref{sec:model_formulation} presents our proposed spatio-temporal-textual point process model as well as model estimation methods. Section~\ref{sec:text-modeling} presents police report text analysis, which will be used in the model as ``marks''. Finally, in Section~\ref{sec:experiments}, we present a study of Atlanta Police data and a comparison with alternative approaches.

\begin{figure}[!ht]
\begin{quote}
\scriptsize{{\bf Call time}: \textit{Oct 12th, 2016, 09:40:00.000}}
\scriptsize{{\bf Location}: \textit{\anonymize{2242 Mt. Paran Rd} Rd NW, Atlanta, GA 30327}}
\scriptsize{{\bf Description}: \textit{I, Ofc. \anonymize{P.Cunningham} assigned to \anonymize{2201} in Vehicle No. \anonymize{30995} was dispatched to \anonymize{2242 Mt. Paran Rd} regarding a \textcolor{red}{residential burglary}. I spoke with the victim, Mrs. \anonymize{Beth Wilkerson} and she advised me that she left her \textcolor{red}{home} around 0940 hrs. this morning and when she returned \textcolor{red}{home} around 1220 hrs. she discovered her side door to her garage \textcolor{red}{kicked in} and the \textcolor{red}{door} to the house was open. Mrs. \anonymize{Wilkerson} went on to say that she had left the door to the house unlocked and did not activate the alarm. Mrs. \anonymize{Wilkerson} advised me that her serving for 12, sterling \textcolor{red}{silver} flatware was taken along with the 2 \textcolor{red}{drawers} that they were in. Mrs. \anonymize{Wilkerson} advised me that the flatware is engraved with ``GFG'', she also stated that the \textcolor{red}{knives} had ``Mother of Pearl handles''. They also stole several (10 or more) sterling \textcolor{red}{silver} serving set pieces and they were also engraved with ``GFG'' or ``D''. she is not sure of the value at this time.  Mrs. \anonymize{Wilkerson} also advised me that her \textcolor{red}{bedroom} was \textcolor{red}{ransacked} and that several pieces of \textcolor{red}{jewelry} are missing. The pieces were \textcolor{red}{diamond} and \textcolor{red}{gold}. Mrs. \anonymize{Wilkerson} advised me that she will have to take an inventory to see what was \textcolor{red}{stolen} and the value. Mrs. \anonymize{Wilkerson} stated that one of the pieces of \textcolor{red}{jewelry} that was \textcolor{red}{stolen} was a \textcolor{red}{gold} and \textcolor{red}{diamond} necklace (Infinity Necklace valued at 8, 000.00 dollars). Mrs. \anonymize{Wilkerson} advised me that the missing \textcolor{red}{jewelry} estimate is around 50, 000 dollars also advised me that after she takes inventory she will make a list and forward it to our CID. I dusted the home for fingerprints with negative results. No cameras or witnesses. I notified Investigator \anonymize{Drew Bahry} of the \textcolor{red}{burglary}.}}
\end{quote}
\caption{A real police report of the residential burglary in Buckhead, Atlanta. Sensitive information has been covered. Note that a set of keywords (in red) is highly correlated with the \emph{M.O.} of this crime. }
\label{fig:police-report}
\end{figure}

\begin{figure}[!h]
\centering
\begin{subfigure}[h]{.4\linewidth}
    \includegraphics[width=\linewidth]{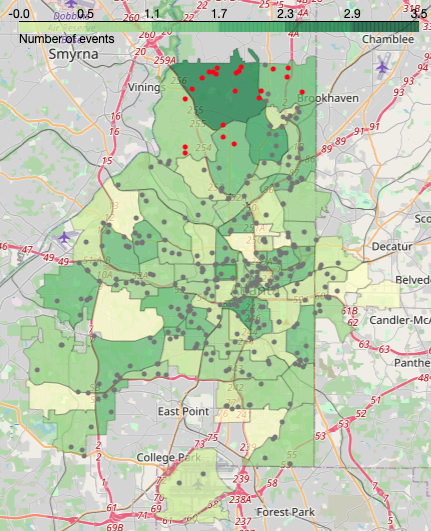}
    \caption{Burglary}
\end{subfigure}
\hspace{.1in}
\begin{subfigure}[h]{.4\linewidth}
    \includegraphics[width=\linewidth]{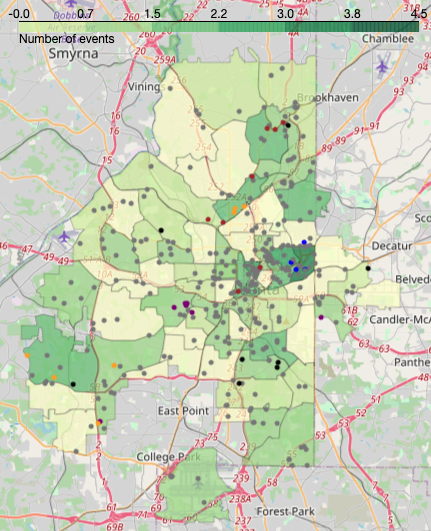}
    \caption{Robbery}
\end{subfigure}
\caption{Spatial distribution of (a) burglary cases and (b) robbery cases in Atlanta from early 2016 to the end of 2017. The red dots in (a) represent an identified series of burglaries committed by the same perpetrator; the orange, blue, brown, black, and purple dots in (b) represent five identified series of robberies; the gray dots represent unidentified burglaries or robberies. The color map indicates the average number of burglaries or robberies reported in each beat. 
We observe that the linked incidents in the same crime series tend to occur within a few (2 to 4) neighboring beats in a relatively small area of the city.}
\label{fig:spatial-dist-exp}
\end{figure}

\subsection{Motivation with real-data example} 
\begin{figure}[!ht]
\centering
\begin{subfigure}[h]{.32\linewidth}
    \includegraphics[width=\linewidth]{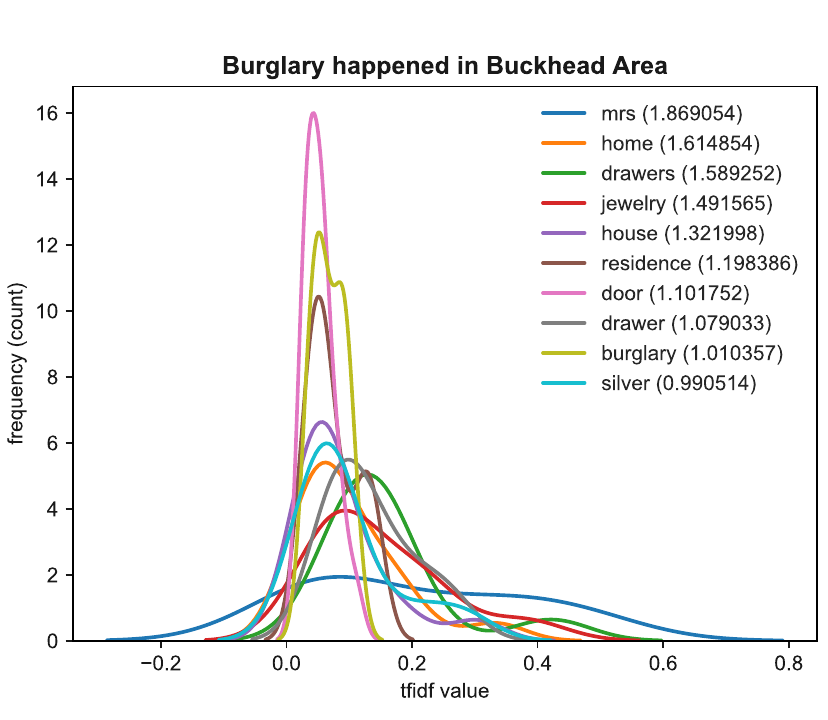}
\end{subfigure}
    \hfill
\begin{subfigure}[h]{.32\linewidth}
    \includegraphics[width=\linewidth]{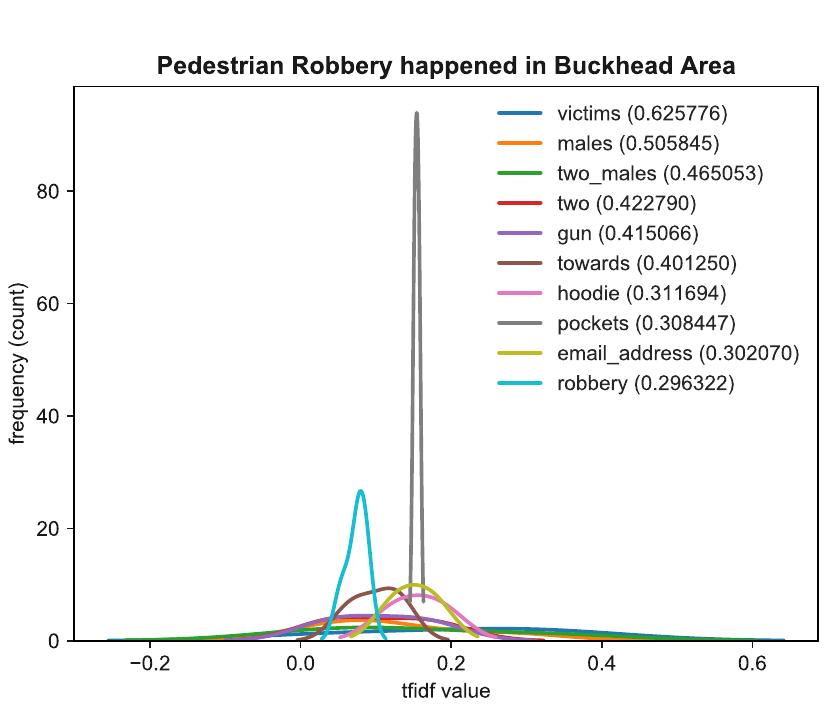}
\end{subfigure}
   \hfill
\begin{subfigure}[h]{.32\linewidth}
    \includegraphics[width=\linewidth]{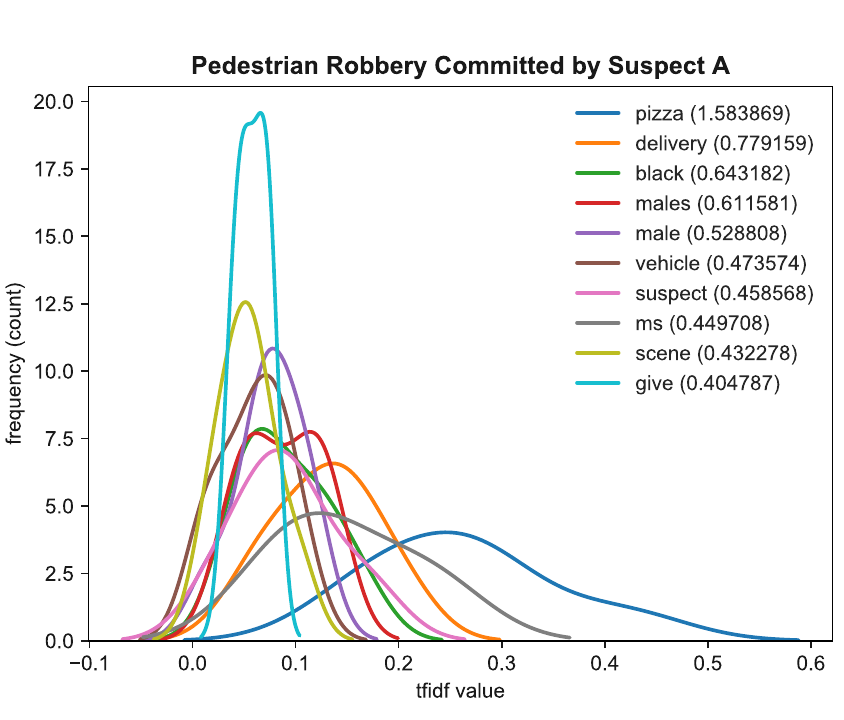}
\end{subfigure}
\vfill
\begin{subfigure}[h]{.32\linewidth}
    \includegraphics[width=\linewidth]{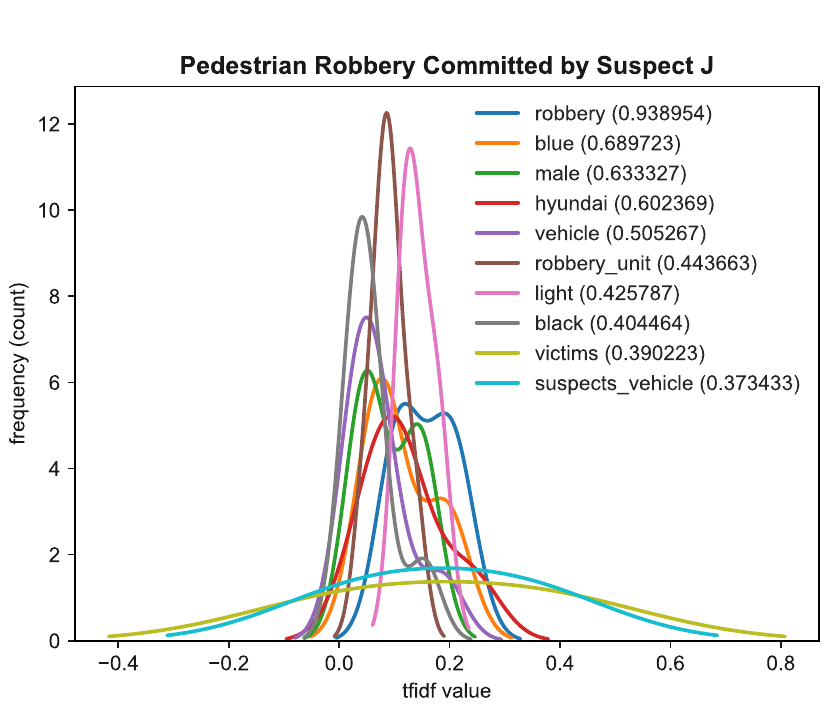}
\end{subfigure}
    \hfill
\begin{subfigure}[h]{.32\linewidth}
    \includegraphics[width=\linewidth]{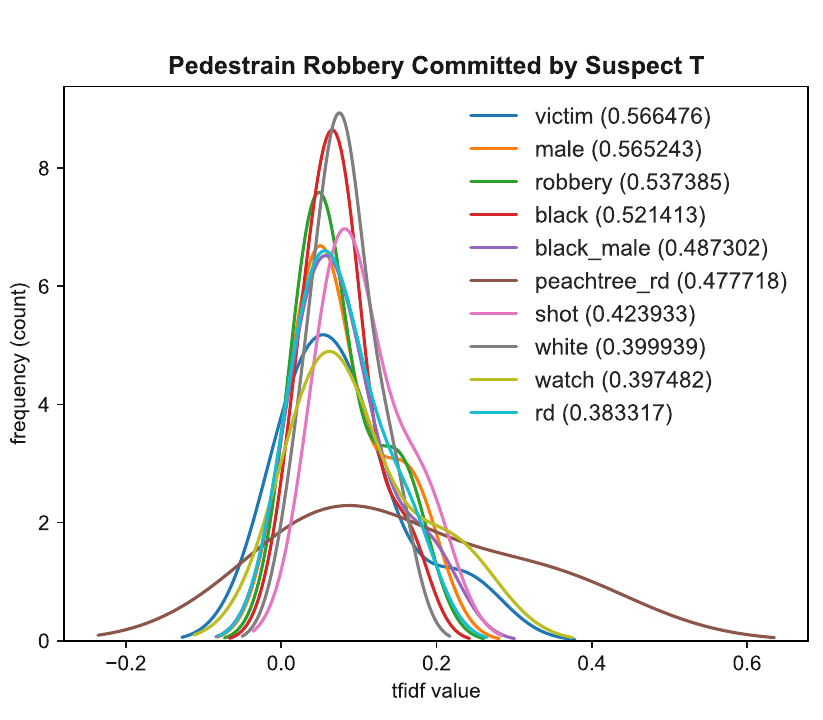}
\end{subfigure}
   \hfill
\begin{subfigure}[h]{.32\linewidth}
    \includegraphics[width=\linewidth]{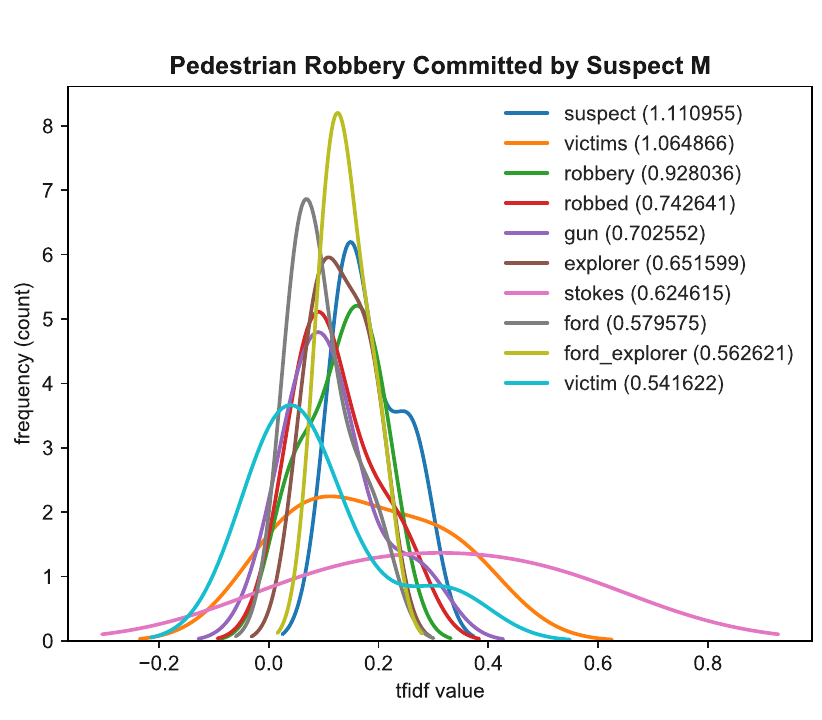}
\end{subfigure}
\caption{Distributions of the top 10 high-frequency keywords for 6 crime series identified by the Atlanta Police. Note that the co-occurring high-frequency keywords for different crime series are different, likely to be related to different \emph{M.O.} of these crime series. 
} 
\label{fig:ngram}
\end{figure}

To motivate the connection in space, time, and police text, first, we present a motivating example. A series of residential burglaries were reported in the Buckhead, a residential neighborhood in Atlanta. From June to November 2016, 23 houses were broken into and stolen. When police arrest the perpetrator, it was found that the same person committed all these burglaries. 
We notice that the incidents in the same crime series tend to aggregate in time and space. This series of burglaries consisting of 23 cases occur within four months, and 12 of them are within just three weeks. Figure~\ref{fig:spatial-dist-exp}a shows the spatial locations of the 23 burglaries, which are clustered within four neighboring beats in a relatively small area. A similar phenomenon can also be observed in another robbery series, shown in Figure~\ref{fig:spatial-dist-exp}b.
Besides the locations and times of the incidents, detailed descriptions have been recorded as text (entered by the police officer who investigated the case). From these police reports, a clear pattern was identified: the affected houses having their bedrooms ransacked, drawers pulled out, and valuable jewelry was stolen. An example of a desensitized police report (with sensitive information masked) on a residential burglary is shown in Figure~\ref{fig:police-report}. Upon examining the document, we notice that the keywords (marked in red) such as \textit{silver}, \textit{bedroom}, \textit{ransacked}, \textit{forced entry}, \textit{bedroom}, \textit{jewelry}, \textit{drawers} frequently appear in the police reports, as shown in Figure~\ref{fig:ngram}. This provides vital clues in detecting crime linkages from many unsolved cases, which is related to the \emph{M.O.} of this crime series. When examining other crime series, we find that a different set of high-frequency keywords occur for different series. Motivated by this, we aim to develop an algorithm that, when combining with time, location information, and the co-occurrence of keywords from police reports, can automatically capture these related incidents and help police investigators to identify M.O. 

\subsection{Related work} According to \cite{Po2016}, there are three main types of approaches to detect crime series, which are \textit{pairwise case linkage}, \textit{reactive linkage}, and \textit{crime series clustering}, respectively. (1) \textit{pairwise linkage detection} [\cite{CoKo2006, LiBr2006, Na2006}] aims to identify whether a pair of crimes were committed by the same offender or criminal group, where each pair is usually considered separately. Such works include \cite{CoKo2006, LiBr2006}, which evaluate the similarity between cases according to the weights determined by experts, and \cite{Na2006}, which learns the similarity from data by considering all incidents jointly. However, they do not consider the common \emph{M.O.} of a crime series. (2) \textit{Reactive linkage} [\cite{WoBuHo2007, Po2016}] is similar to \textit{pairwise case linkage}, which starts with a seed of one or more crimes, and discovers one crime at a time for a crime series. (3) \textit{Crime series clustering} [\cite{Po2016, Ad2004, AdMu2003, DaMu2003, MaChHu2010, WaRuWaSe2015}] discovers all clusters of crime incidents simultaneously; however, this approach requires labels, which is infeasible in practice. 

Recently, there has been much work on modeling discrete incident data using point processes. In particular, Hawkes processes (also known as self-exciting point processes) have been widely used for studying human dynamics [\cite{mohler2011self, lai2016topic, fox2016spatially}], which are characterized by the mutual ``triggering'' effect among incidents. Such models can capture inhomogeneous inter-incident times and causal (spatial and temporal) effect. Our method extends the vanilla version of multivariate Hawkes processes by considering the unstructured text information as marks of the incidents. Unlike the stochastic declustering methods developed by \cite{zhuang2002stochastic, zhuang2004analyzing, veen2008estimation}, which require an exact parametric form of triggering function, we derive Expectation Maximization (EM) algorithm [\cite{dempster1977maximum, mclachlan2007algorithm}] to estimate our model.

There are works studying general correlations between incidents that are not necessarily crime incidents by performing incident embeddings and evaluating their similarities in the embedding space. Such works include \cite{ZhLiLeYuZhHaHa2017}, which uses tweet token as context while capturing correlation between time, location, and keywords (tokens) in the same tweet; \cite{HoWuLiWu2017, DuDaTrUpGoSo2016}, which are based on Recurrent Neural Networks (RNN) and treat category as marks of incidents; \cite{Quinn2011}, which infers the causal relationship for neural data. Our problem's particular challenge is that our data involves complex marks, which are the police report's text description. 

The \textit{spatio-temporal-textual} incident data has also been consider in \cite{LiJiLu2010, AnRoCo2017, WaDoBoLuYuChDa2012}. However, these existing works do not use Hawkes process models nor consider crime linkage detection. Another related work [\cite{KuBrBe2017}] uses the topic model for text, and it solves the problem of crime category classification, which is very different in nature from crime linkage detection.

Existing works on regularized RBMs consider various forms of regularization on the weights of RBMs to yield sparse outputs [\cite{Luo2011, Halkias2013, Keyvanrad2017, Shen2019}] or sparse feature [\cite{Ranzato2006, Ranzato2007}]. However, they mainly focus on producing sparsity in the neural network connection, which is different from the probabilistic sparsity required for the keywords selection in our scenario. Our paper extends our prior preliminary work [\cite{ZhXiRBM2018, ZhXiUFS2018}], which only considers text information. 


\section{Model}
\label{sec:model_formulation}

Consider a sequence of $n$ \textit{spatio-temporal-textual} incidents, where each observation is a tuple consisting of time, location, and text:
\begin{equation}
(t_1, s_1, \boldsymbol{x}_1), (t_2, s_2, \boldsymbol{x}_2), \dots, (t_n, s_n, \boldsymbol{x}_n).
\label{model}
\end{equation} For the $i$th incident: $t_i \in [0, T]$ denotes time, $T$ is the time horizon, and $t_i < t_{i+1}$; $s_i \in \mathcal{S} \subset \mathbb{R}^2$ denotes the spatial location of the $i$th incident that consists of the latitude and longitude of the incident;
$\boldsymbol{x}_i = [x_{i1}, x_{i2}, \dots, x_{iq}]^\intercal \in \mathbb{R}^q$ corresponds to the fixed-length \textit{Bag-of-Words} representation [\cite{Ze1954}] with $q$ keywords. 

Here $x_{ij}$ is the \textit{TF-IDF} (Term-Frequency-Inverse-Document-Frequency) value [\cite{WaAl2013}] of the $j$th keyword and $q$ is the total number of the keywords that appeared in the corpus (the collection of all text documents). We will design a mapping function $\varphi: \mathbb R^q \to \{0, 1\}^m$ (further discussed in Section~\ref{sec:text-modeling}) to project the \textit{Bag-of-Words} representation into the $m$-bit binary embeddings $\boldsymbol{h} \in \{0, 1\}^m$: $\boldsymbol{h} =\varphi(\boldsymbol{x})$. 

\subsection{Spatio-temporal-textual process}

We model the \textit{spatio-temporal-textual} incident data using multivariate marked Hawkes processes [\cite{DaVe2007}]. 
A widely accepted hypothesis is that crime incidents exhibit the so-called triggering effect: once a crime incident occurs, this will increase the chance of similar incidents in the nearby neighborhood and the near future [\cite{mohler2011self, mohler2014marked, park2021investigating}]. This effect motivates the adoption of the self-exciting Hawkes processes. 
Let $\mathcal{H}_t$ denote the $\sigma$-algebra generated by all historical incidents before time $t$. 
Therefore, the conditional intensity function of the Hawkes process [\cite{Ra2011}] defines the probability density that an incident occurs at the location $s$, at time $t$, and with text $\boldsymbol{h}$, conditioning on the history $\mathcal{H}_t$ of incidents happens before time $t$: 
\begin{equation}
\lambda(s, t, \boldsymbol{h} | \mathcal{H}_t) = \underset{\Delta s, \Delta \boldsymbol{h}, \Delta t \to 0}{\lim} \frac{\mathbb{E}\left[N(B(s, \Delta s) \times B(\boldsymbol{h}, \Delta \boldsymbol{h}) \times [t, t + \Delta t))\right|\mathcal{H}_t]}{\left | B(s, \Delta s)\right | \left | B(\boldsymbol{h}, \Delta \boldsymbol{h}) \right | \Delta t},
\end{equation}
where $N(C)$ is the counting measure defined as the number of incidents that occur in the set $C \subseteq [0, T] \times \mathcal{S} \times \mathbb{R}^{m}$, and $\left | B(v, \Delta v) \right |$ denotes the Lebesgue measure of a ball centered at $v$ with the radius $\Delta v$.
Hawkes process is a \textit{self-exciting} point process with the conditional intensity being influenced by the past incidents positively: 
\begin{equation}
\lambda(s, t, \boldsymbol{h}|\mathcal{H}_t) = \mu(s) + \sum_{j:t_j<t} g(s, s_j, t, t_j, \boldsymbol{h}, \boldsymbol{h}_j),
\label{eq:con_intensity1}
\end{equation}
where $\mu(s)$ is the base intensity.
Below, we omit $\mathcal{H}_t$ from the notation while remembering  it is a conditional intensity.

We assume the triggering kernel function to be separable in space, time, and marks, as commonly assumed in the point process literature (see the review in [\cite{Re2018}]): 
\[
g(s, s_j, t, t_j, \boldsymbol{h}, \boldsymbol{h}_i) = 
\upsilon(s,s_j) \nu(t,t_j) \kappa(\boldsymbol{h}, \boldsymbol{h}_j) 
\geq 0,
\] 
where $\upsilon, \nu, \kappa$ are three kernel functions for two incidents in time, location, and mark spaces, respectively.
We would like to remark that the separable kernel will lead to a more computationally efficient procedure since the evaluation of the likelihood function requires integrating the intensity function over the whole space, which is high-dimensional, as explained later in Section \ref{sec:likelihood}. Moreover, separable kernels in our case lead to interpretable results. To achieve this goal, we made the following choices for the kernels in our paper out of many possible forms. (1) For the temporal kernel
$\nu$, we assume a commonly used exponential function $\nu(t, t_j) = \beta \exp\{-\beta (t-t_j)\}$, where $t > t_j$ and the parameter $\beta > 0$ captures the decay rate of the influence, since linked crime incidents usually aggregate in time; note that the kernel integrates to one over $t$. (2) Since police departments operate by \textit{beats} (geographical units for police patrolling), we discretize the location into $d$ disjoint units (according to beats), and replace the location $s$ by the beat index $k \in \{1,2,\dots, d\}$. After discretization, the spatial function $\upsilon(s, s_j)$ is represented by the coefficient $\alpha_{s, s_j}$, the influence strength of beat $s_j$ to beat $s$. If $\alpha_{u, v} = 0$, then beat $v$ has no influence to beat $u$. 
Note that the spatial influence can be directional, i.e., $\alpha_{u, v}\neq \alpha_{v, u}$. Define the coefficient matrix $A=\{\alpha_{u,v}\} \in \mathbb R^{d \times d}, \alpha_{u,v} \ge 0$.
(3) For text, we choose the inner product between text embeddings as kernel function. 
Since {\it textual similarity} is commonly measured by the normalized inner product between embeddings, we use this as our kernel function
$\kappa(\boldsymbol{h}, \boldsymbol{h}_j)=\boldsymbol{h}^\intercal \boldsymbol{h}_j/m := \tilde{\boldsymbol{h}}^\intercal \tilde{\boldsymbol{h}}_j$, where $\tilde{\boldsymbol{h}}$ and $\tilde{\boldsymbol{h}}_j$ denote normalized embeddings. 

The rationale for choosing the form of the intensity function as above is tri-fold. (1) The influence of incidents is causal: the current incident only depends on past incidents, and their influence decays over time. (2) The spatial coefficient measures the correlation between two discrete locations, and the correlation may not decay over the distance since, in our context, crime incidents are not necessarily linked to what happens in the nearby neighborhood (e.g., criminals may travel). This approach allows us to capture more complex spatial influence. (3) We assume two incidents with higher textual similarity are more likely to be linked since text similarity can imply similar (\emph{M.O.}). Moreover, we choose to use the inner product of embedding because it is the most common measure in natural language processing. Also, in our setting, this leads to the closed-form likelihood function.

Following the above modeling assumptions, the conditional intensity of the $k$th dimension can be written as
\begin{equation}
{\lambda}_t^k(\boldsymbol{h}) = 
\mu_k + \sum_{j:t_j < t} \alpha_{k, s_j} \beta e^{-\beta (t - t_j)} 
\tilde{\boldsymbol{h}}^\intercal \tilde{\boldsymbol{h}}_j, 
\quad \forall t, k,
\label{eq:cond_intensity3}
\end{equation}
where $\mu_k$ is a base intensity for beat $k$, 
which can be related to the ambient crime rate of the beat. Note that we use an unconventional approach to model marks by including them as a part of the kernel, which can model the triggering effect of the incidents with similar $\boldsymbol{h}$, as motivated by the fact that crime incidents with similar \emph{M.O.} tend to trigger each other. In contrast, the conventional marked point process usually assumes the intensity function to be factorized into two terms: the conditional intensity function $\lambda_g(s, t)$ of the space and time (independent of marks), and the conditional distribution $f(x | k, t)$ of the mark given time and location:
$
    \lambda(k, t, \boldsymbol{h}) = \lambda_g(k, t) f(\boldsymbol{h} | k, t).
$



\subsection{Model estimation and likelihood function}\label{sec:likelihood}
\label{sec:learning}

In this section, we explain the estimation procedure of our model. First, we choose to estimate the base intensity $\{\mu_k\}$ by the average number of incidents that occurred in that beat, which can be viewed as a non-parametric estimation procedure; similar non-parametric estimation for the base intensity using kernel estimation has been considered in [\cite{mohler2014marked}]. Our approach is
computationally more efficient compared to the traditional stochastic declustering algorithm. We also validate that the performance of our approach is comparable to that of the stochastic declustering in Appendix~\ref{append:background}.

The spatial coefficient $A$ are estimated based on the likelihood function. The log-likelihood function for $n$ incidents in $[0, T]$ can be derived based on the conditional intensity \eqref{eq:cond_intensity3} (detailed derivation in Appendix~\ref{append:sttpp-likelihood}):
\begin{equation}
\begin{aligned}
\ell (A)
= &~\sum_{i=1}^{n} \log \left ( \mu_{s_i} + \sum_{j=1}^{i-1} \alpha_{s_i, s_j} \beta e^{- \beta (t_i -t_j)} \tilde{\boldsymbol{h}}_i^\intercal \tilde{\boldsymbol{h}}_j \right ) \\
&~- \sum_{k=1}^{d} \mu_k |\Omega| T - \sum_{j=1}^n \sum_{k=1}^{d} \sum_{\boldsymbol{h} \in \Omega} \alpha_{k, s_j} \left (1 - e^{-\beta(T - t_j)} \right ) \tilde{\boldsymbol{h}}^\intercal \tilde{\boldsymbol{h}}_j.
\end{aligned}
\label{eq:pp_log_likelihood}
\end{equation}
Given fixed $\beta$, the spatial coefficient can be estimated by maximum likelihood $\widehat A:=\arg\max_A \ell(A)$; the related optimization problem can be solved efficiently by an Expectation-Maximization (EM) algorithm in Section \ref{sec:EM}.
It can be shown that $\ell(A)$ is concave [\cite{SiJo2010}] and, hence, there is a unique global maximizer. 

We treat the influence parameter $\beta >0$ as a tuning parameter that is estimated separately (discussed in Section \ref{sec:temporal_coefficient}). We take this approach because if we treat both $A$ and $\beta$ as unknown, the corresponding maximum likelihood problem is non-convex and we cannot easily find a global optimal solution. 

\subsection{Spatial coefficients and crime linkage estimation by EM algorithm}\label{sec:EM} 

In this section, we discuss the model estimation procedure based on the Expectation-Maximization (EM) algorithm, and also explains how to evaluate the likelihood that two crime incidents are linked by introducing a set of auxiliary variables. The EM algorithm is derived following the similar strategy as in \cite{Re2018}.  Introduce a set of auxiliary variables $\{p_{ij}\}$ satisfying \[
\forall i,\ \sum_{j=1}^i p_{ij} = 1,\ p_{ij} \ge 0, 
\]
where $\{p_{ij}\}, \forall i, j : i > j$ can be interpreted as the probability that the $i$th incident is triggered by the $j$th incident (i.e., there is a linkage between $i$th and $j$th incidents), and $\{p_{ii}\}, \forall i$ can be interpreted as the probability that the $i$th incident is generated due to background process. These auxiliary variables will enable us to derive a computationally efficient EM algorithm, and they can also be used for crime linkage detection: given a incident of interest $i$, we can use $p_{ij}\in[0,1]$ to measure the likelihood that $j$ is triggered by $i$ as explained below. As shown in Appendix~\ref{append:lower-bound-log-likehood}, we can obtain a lower bound to the likelihood function \eqref{eq:pp_log_likelihood} using Jensen's inequality:
\begin{equation}
\begin{aligned} 
\ell(A) \ge &~\sum_{i=1}^n \Biggl( p_{ii}\log(\mu_{s_i}) + \sum_{j=1}^{i-1} p_{ij} \log \left( \alpha_{s_i, s_j} \beta e^{-\beta (t_i - t_j)} \tilde{\boldsymbol{h}}_i^\intercal \tilde{\boldsymbol{h}}_j \right) \\
&~- \sum_{j=1}^i p_{ij} \log p_{ij} \Biggr) - \sum_{k=1}^d \mu_k |\Omega| T - \sum_{k=1}^{d} \sum_{j=1}^n \sum_{\boldsymbol{h} \in \Omega} \alpha_{k,s_j} \left( 1 - e^{-\beta (T - t_j)} \right) \tilde{\boldsymbol{h}}^\intercal \tilde{\boldsymbol{h}}_j.
\label{eq:pp_log_likelihood_lowerb}
\end{aligned}
\end{equation}
From the form of the lower bound, it can be seen that the $\{p_{ij}\}$ can be interpreted as the probability that one incident triggers another.
When maximizing the lower bound, the optimizers can be expressed in closed-forms as shown in Appendix~\ref{append:grads-em}. Thus, the EM algorithm involves the following iteration: in each iteration $r$,
\begin{subequations}
\begin{align}
p_{ii}^{(r)} & = \frac{\mu_{s_i}}{\mu_{s_i} + \sum_{l=1}^{i-1} \alpha_{s_i, s_l}^{(r)} \beta e^{-\beta (t_i - t_l)} \tilde{\boldsymbol{h}}_i^\intercal \tilde{\boldsymbol{h}}_l }, \label{eq:estep-1} \\
p_{ij}^{(r)} & = \frac{\alpha_{s_i, s_j}^{(r)} \beta e^{-\beta(t_i - t_j)} \tilde{\boldsymbol{h}}_i^\intercal \tilde{\boldsymbol{h}}_j }{\mu_{s_i} + \sum_{l=1}^{i-1} \alpha_{s_i, s_l}^{(r)} \beta e^{-\beta (t_i - t_l)} \tilde{\boldsymbol{h}}_i^\intercal \tilde{\boldsymbol{h}}_l }, ~j < i, \label{eq:estep-2} \\
\alpha_{u, v}^{(r+1)} & = \frac{\sum_{i=1}^{n} \sum_{j=1}^{i-1} \mathbb{I}\{s_i = u, s_j = v\} p_{ij} }{\sum_{j=1}^{n} \mathbb{I}\{s_j = v\} (1 - e^{-\beta (T - t_j)}) \sum_{\boldsymbol{h} \in \Omega} \tilde{\boldsymbol{h}}^\intercal \tilde{\boldsymbol{h}}_j}. \label{eq:mstep}
\end{align}
\end{subequations}

Given the pre-computed text embeddings, the EM algorithm can be performed efficiently. Moreover, 
when implementing the algorithm, it appears that we need to sum $\boldsymbol{h} \in \Omega$; in fact, we can simplify the computation while achieving good performance. Rather than naively enumerating all possible embeddings, which will result in summing over $2^m$ terms, we first perform text embedding (discussed in the next section) and examine the actual support of the learned embeddings $\boldsymbol{h}$. Then we define $\Omega$ as the union of the observed embeddings from training data. The resulted $|\Omega| \ll 2^m$. For instance, for our real-data corpus with 10,056 documents, the embedding uses $m = 1000$, and a naive enumeration will require to sum $2^{1000}$ terms. Using our simplification, the size of the set $|\Omega|$ is 1,743. Based on the result of the EM algorithm, finding the most related incidents of the $i$th incident can be done by selecting incidents with the largest $p_{ij}$. 

\section{Text embedding with keyword selection}
\label{sec:text-modeling}

In this section, we present our text embedding method with keywords selection. 
Recall that we treat the text embedding as marks for the spatio-temporal Hawkes process. 
The reason for considering text embedding $\boldsymbol{h}$ for marks instead of the raw Bag-of-Words representation $\boldsymbol{x}$ is two-fold: (1) The EM algorithm (Equations \ref{eq:estep-1}-\ref{eq:mstep}), if we were to use $\boldsymbol{x}$ (a 7,039 dimensional real-valued vector), would require to integrate $\boldsymbol{x}$ over the entire space $\Omega$; however, high-dimensional numerical integral is intractable in general. Using the embedding representation, in the EM algorithm, we need to sum over $\boldsymbol{h}$ (a 1000 dimensional binary vector), which is a much simpler task than integration over $\boldsymbol{x}$. (2) The police report text is unstructured and contains much noisy information; text embedding can be viewed as a certain ``denoising'' procedure. Moreover, the important information about the crime incidents is contained in the frequency of certain keywords that appeared in the entire corpus, as illustrated in Figure~\ref{fig:police-report}; embedding can capture the inherent latent connection from the raw text.
Note that we perform the text embedding and model estimation for the Hawkes processes separately because jointly learning the two is mathematically intractable.

Text embedding is a commonly used technique in natural language processing [\cite{MiSuChCoDe2013}], where each document is viewed as a combination of a set of keywords. The idea is to map words with similar semantic meanings to be closer to each other in the embedding space. Here we represent each police report as a feature vector using the \textit{Bag-of-Words} representation. To perform embedding, we use the Restricted Boltzmann Machine (RBM) [\cite{FiIg2012}], which characterizes the joint distribution of keywords and unknown latent variables (the embeddings). Moreover, we introduce a new regularization function for keyword selection, which is important for detecting crime linkages. Recall the example in Figure~\ref{fig:ngram}, where documents in different crime series tend to have a different distribution of high-frequency keywords. Thus, these co-occurrent keywords in each crime series tend to be highly related to the \emph{M.O.} of the crime series.  Since such high keywords defining \emph{M.O.} are only a small portion of the entire vocabulary, it motivates us to perform keyword selection in embedding based on a proper regularization. Thus, we penalize the total probability that the keywords are ``active'' in the model. 


\begin{figure}
\centering
\includegraphics[width=1.\linewidth]{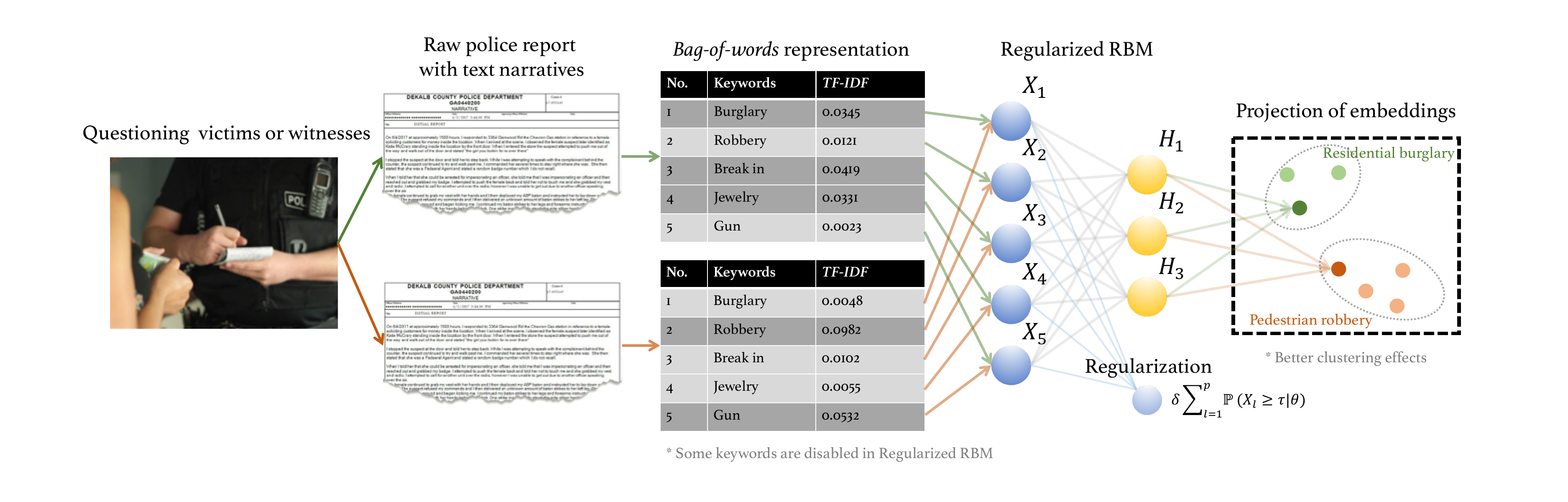}
\caption{Structure of the regularized RBM for text embedding.}
\label{fig:reg-rbm}
\end{figure}

\subsection{Restricted Boltzmann Machine (RBM)}

An RBM is a probability graphical model, which can also be viewed as a two-layer neural network. As shown in Figure~\ref{fig:reg-rbm}, the RBM consists of two types of units, the so-called visible and hidden units. In our context, the visible units correspond to keywords, and the hidden units are also called the embeddings.  Assume that the visible layer has $q$ units, denoted by a vector $\boldsymbol{X} = [X_1, X_2, \dots, X_q]^\intercal \in \mathbb{R}^q$, and the hidden layer has $m$ units, denoted by $\boldsymbol{H} = [H_1, H_2, \dots, H_m]^\intercal \in \{0, 1\}^m$. Parameters of the network are $\theta = \{\boldsymbol{w}, \boldsymbol{b}, \boldsymbol{c}\}$, which include weights $\boldsymbol{w} = \{w_{lj}\} \in \mathbb{R}^{q \times m}$, visible bias $\boldsymbol{b} = \{b_l\} \in \mathbb{R}^{q}$, and hidden bias $\boldsymbol{c} = \{c_j\} \in \mathbb{R}^{m}$. Given training data, the parameters of RBM are tuned to maximize the likelihood of the data under the model. 

Consider the Gaussian-Bernoulli RBM, where the joint distribution of visible and hidden units are specified by: 
\[p(\boldsymbol{X}, \boldsymbol{H} | \theta) = \frac 1 Z \exp\{-E_\theta(\boldsymbol{X},\boldsymbol{H})\},\] and the partition function $Z = \sum_{\boldsymbol{X}, \boldsymbol{H}} e^{-E_\theta(\boldsymbol{X}, \boldsymbol{H})}$ is a normalization constant. The energy function is given by 
\begin{equation}
E_{\theta}(\boldsymbol{X}, \boldsymbol{H}) = \sum\nolimits_{l=1}^{q} \frac{(X_l - b_l)^2} {2 \sigma^2} - \sum\nolimits_{j=1}^{m} c_j H_j - \sum\nolimits_{l=1}^{q} \sum\nolimits_{j=1}^{m}\frac{X_l}{\sigma} H_j w_{lj},
\end{equation}
where $\sigma^2$ is the variance of the Gaussian noise for visible variables. It can be shown that the visible and hidden variables are independent upon conditioning on the hidden and the visible variables, respectively:
$p(\boldsymbol{X}|\boldsymbol{H};\theta) = \prod_{l=1}^q p(X_l | \boldsymbol{H}; \theta)$, and 
$p(\boldsymbol{H}|\boldsymbol{X};\theta) = \prod_{j=1}^m p(H_j | \boldsymbol{X}; \theta).
$ This property simplifies our derivation. Let $\mathcal N(x; \mu, \sigma^2)$ denote the probability density function of normal random variable with mean $\mu$ and variance $\sigma^2$. 
The conditional probabilities for the $i$th keyword and the $j$th entry in the embedding are given by
\begin{subequations}
\begin{align}
  p(X_l|\boldsymbol{H};\theta) &= \mathcal N\left(X_l; b_l + \sigma\sum\nolimits_{j=1}^m w_{lj}H_j, \sigma^2\right), \label{eq:cond-prob-visible}\\
  p(H_j = 1|\boldsymbol{X};\theta) &= \text{sigm}\left(c_j +\sum\nolimits_{l=1}^q \frac{X_l}{\sigma} w_{ij} \right), \label{eq:cond-prob-hidden}
\end{align}
\end{subequations}
where the sigmoid function is defined as $\text{sigm}(x) = 1 / (1 + e^{-x})$. The marginal distribution of keywords is given by
\[p(\boldsymbol{X} | \theta) = \sum\nolimits_{\boldsymbol{H}} p(\boldsymbol{X}, \boldsymbol{H} | \theta) = \frac 1 Z \sum\nolimits_{\boldsymbol{H}} \exp\{-E_{\theta}(\boldsymbol{X}, \boldsymbol{H})\} ,\] which is also known as the Gibbs distribution. 

\subsection{Regularized RBM}

Consider independent and identically distributed (i.i.d.) training data $\boldsymbol{x}_1, \ldots, \boldsymbol{x}_n$. The RBM's log-likelihood function is given by 
$
\mathcal{L}(\theta) = \sum_{i=1}^n \log p(\boldsymbol{x}_i | \theta).
$
We now present a regularized maximum likelihood estimate of the RBM model:
\begin{equation}
{\min}_\theta \Bigl \{-\mathcal{L}(\theta) + \delta \sum\nolimits_{l = 1}^{q} \mathbb{P}(X_l\ge \tau |\theta) \Bigr \},
\label{eq:new_obj}
\end{equation}
where $\delta > 0$ is the regularization parameter, and the threshold $\tau$ a hyper-parameter that controls the threshold for keywords selection. We assume that keywords with TF-IDF value less than $10^{-2}$ can be ignored and set $\tau = 10^{-2}$ in experiments. The regularization function penalizes the total probability of visible units (keywords) being ``selected'' by the model, and thus, it encourages ``stochastic sparse'' activation patterns and only selects a subset of keywords. Here we do not directly use the standard $\ell_1$-norm type of regularizer on the weights because we want the output to be sparse in a probabilistic sense.


Computing the likelihood of a Markov random field or its gradient is, usually, computationally intensive. Thus, we employ sampling-based methods to approximate the likelihood function and its gradient [\cite{hinton2012practical}] and perform stochastic gradient descent [\cite{lan2020first}]. In each iteration, we optimize one variable while fixing other variables, and gradients are evaluated from samples. 

A benefit of our regularization function is that its gradient can be derived in closed-form. Below, let $\phi(\cdot)$ and $\Phi(\cdot)$ denote the probability density function and cumulative distribution function of the standard normal random variable, respectively. Let $\left < \cdot \right >_P$ denote the expectation with respect to a distribution $P$. 
We can write the regularization term as
\begin{equation}
\mathbb{P}(X_l\ge \tau |\theta)
= \langle 
\mathbb{P}(X_l \ge \tau | \boldsymbol{H}; \theta)\rangle_{p(\boldsymbol{H})} 
= 1-\left \langle \Phi
\left(\frac{\tau - b_l - \sigma \sum\nolimits_{j=1}^{m} w_{lj} H_j}{\sigma}\right)\right \rangle_{p(\boldsymbol{H})}.
\end{equation} 
Let \[\tau_l'=\tau - b_l- \sigma \sum_{j=1}^m H_j w_{lj}, \quad l = 1, \ldots, n.\] 
The detailed derivation of gradients is shown in Appendix~\ref{append:reg-rbm-loglikelihood}. This leads to a simple procedure for performing stochastic gradient descent in the parameters of the RBM model: 
\begin{align*}
\Delta w_{lj} = & ~
\left < X_l H_j\right >_{p(\boldsymbol{H}|\boldsymbol{X})\tilde{p}(\boldsymbol{X})}  - \left < X_l H_j\right >_{p(\boldsymbol{X}, \boldsymbol{H})} - \delta \left\langle 
 \frac{H_j \phi(\tau_l')}{1-\Phi(\tau_l')} 
\right\rangle_{p(\boldsymbol{H}|\boldsymbol{X})\tilde{p}(\boldsymbol{X})}, \\
\Delta b_{l} = & ~
X_l - \left < X_l \right >_{p(\boldsymbol{X})} 
-   \frac{\delta}{2\sigma^2}\ \left\langle 
 \frac{\phi(\tau_l')}{1-\Phi(\tau_l')} 
\right\rangle_{p(\boldsymbol{H}|\boldsymbol{X})\tilde{p}(\boldsymbol{X})},\\
\Delta c_{j} = & ~
p(H_j=1|\boldsymbol{X}) - \left < p(H_j=1|\boldsymbol{X})\right >_{p(\boldsymbol{X})},
\end{align*}
where $\tilde{p}(\boldsymbol{X})$ denotes the empirical distribution. To evaluate the gradient, we adopt the $k$-step contrastive divergence (CD-$k$) algorithm [\cite{Hi2002}].

\section{Real-data study}
\label{sec:experiments}

We now study a large-scale police dataset using our methods and demonstrate its competitive performance for linkage detection.

\subsection{Dataset}
\label{sec:dataset}

We study a data set of 10,056 crime incidents recorded by the Atlanta Police Department from early 2016 to the end of 2017. Each incident is associated with a 911 call, with information including crime category, time and location of the incident, and comprehensive text descriptions entered by the police officer. Two crime incidents are said to be linked if they are in the same crime series. We only have a handful of identified crime series by police, consisting of 6 crime series and a total of 56 incidents in these series. Thus, this also shows the importance of an unsupervised learning approach. Here, the labels for the crime series are not used for fitting the model but are only used for validation. 

We preprocess the raw data as follows.
(1) \textit{Discretize the continuous geolocation of the crime incidents according to beats}. We associate each crime incident using the policing beat index. Atlanta is divided into 80 disjoint beats. 
(2) \textit{Normalize the call time}. We regard the call time (when the dispatch center receives the 911 call) as each crime incident's initial time. For ease of calculation, we then normalize this time to the range of $[0, 1]$, i.e., the time horizon corresponds to $T = 1$. 
(3) \textit{Initialize base intensities}. We estimate the base intensity by estimating the average number of incidents in each beat and within the time horizon. 
(4) \textit{Construct Bag-of-Words representations for text documents}. We normalize the text to lower-cases so that, e.g., the distinction between ``The'' and ``the'' are ignored; we also remove stop-words, independent punctuation, low term-frequency (TF) terms, and the terms that appeared in most documents (high document frequency terms). Then we compute the \textit{Bag-of-Words} vector for each police report using 7,039 keywords. For the document's feature vector, each entry corresponds to the \textit{TF-IDF} value of a keyword or bi-gram (a sequence of two adjacent words from a string of text) that appears in the corpus. 
(5) We study two particular categories of crime: burglary and robbery, as the same category cases may define similar \emph{M.O.} There are 349 \emph{burglary} crimes (23 of them are labeled) and 333 \emph{robbery} crimes (23 of them are labeled), respectively. We also compare with groups with 305 \emph{mixed} types of crimes (56 of them are labeled). 

\subsection{Text embedding results}

\begin{figure}[!t]
\vspace{-.1in}
\centering
\includegraphics[width=.4\linewidth]{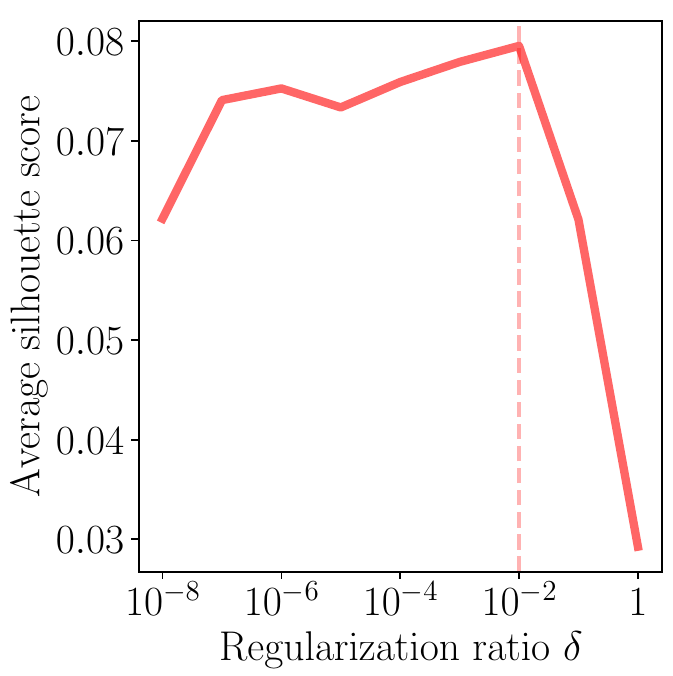}
\caption{Optimal choice of regularization parameter $\delta$ for regularized RBM by $5$-fold cross-validation. The vertical axis corresponds to average silhouette scores for clustering ``burglary'', ``robbery'', and ``mixed'' types of incidents in the embedding space by regularized RBM. The model attains its best performance at $\delta = 10^{-2}$, which we use in the subsequent experiments.
}
\vspace{-.1in}
\label{fig:cv-delta}
\end{figure}
 
First, we examine the text embedding and keywords selection results.  Recall that regularized RBM aims to select the most relevant keywords for clustering in the embedding space. The number of keywords selected in the model depends on the regularization parameter value, which is typically chosen by cross-validation. Since we have very few labeled crime series data, we will take the following approach for cross-validation. Consider two types of police reports, including ``burglary'', ``robbery'', and group the rest of the category as ``mixed'' type. Based on this, we examine the embedding space to map the police reports to the same category as much as possible (use this approach as weak supervision). Specifically, we use the average \emph{silhouette score} [\cite{Rousseeuw1987}] as a metric, which measures the average distance of a point to its cluster (cohesion) relative to other clusters (separation). The silhouette ranges from -1 to 1 (the higher, the better), and we select the value of the regularizing parameter $\delta$ to maximize the average silhouette score. As shown in Figure~\ref{fig:cv-delta}, based on $5$-fold cross-validation, the optimal choice of $\delta=10^{-2}$.

\begin{figure}[!tt]
\vspace{-.1in}
\centering
\includegraphics[width=0.8\linewidth]{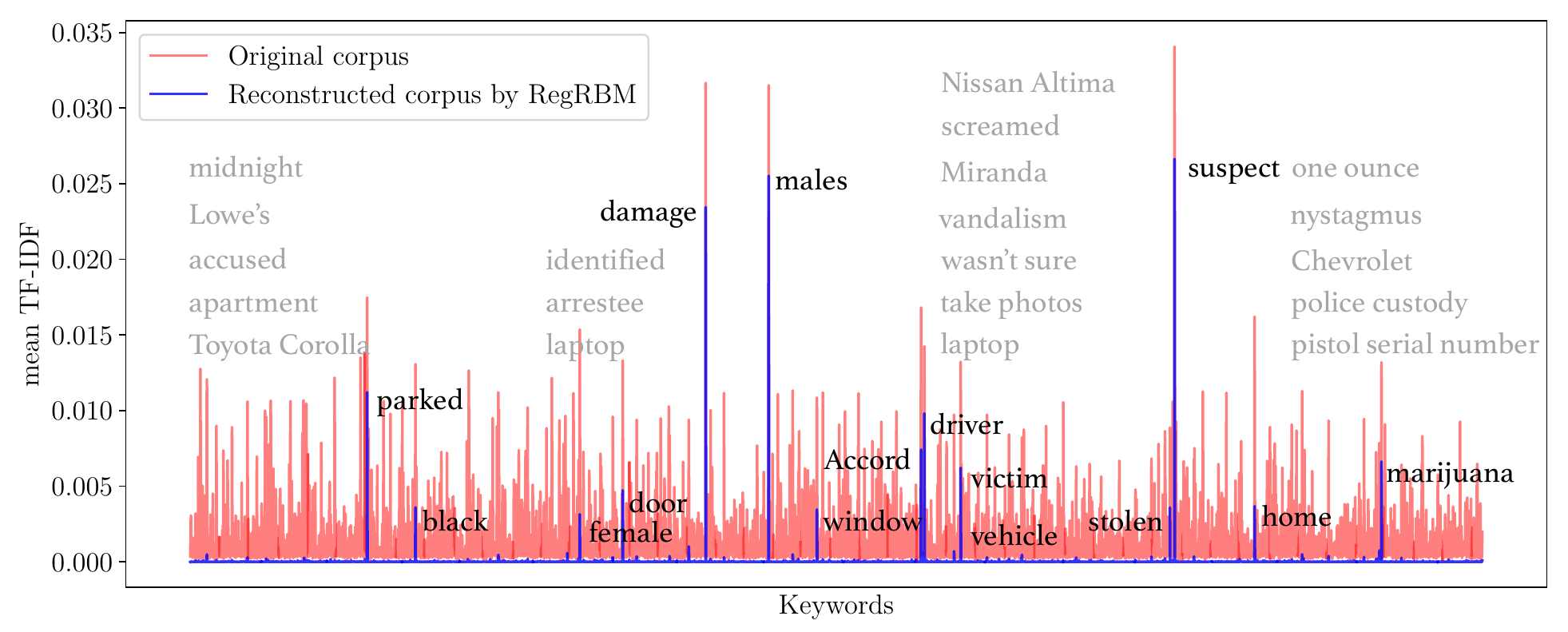}
\vspace{-.1in}
\caption{The average \textit{TF-IDF} value of keywords in the original corpus (shown in red) and the selected keywords (shown in blue). Crime analysts in the Atlanta Police Department validate that the selected keywords by our model, such as \textit{stolen}, \textit{home}, \textit{marijuana},  play a significant role in defining the \emph{M.O.} and linking crime incidents.}
\label{fig:corpus}
\end{figure}

\begin{figure}[h!]
\centering
\begin{subfigure}[h]{0.48\linewidth}
\includegraphics[width=\linewidth]{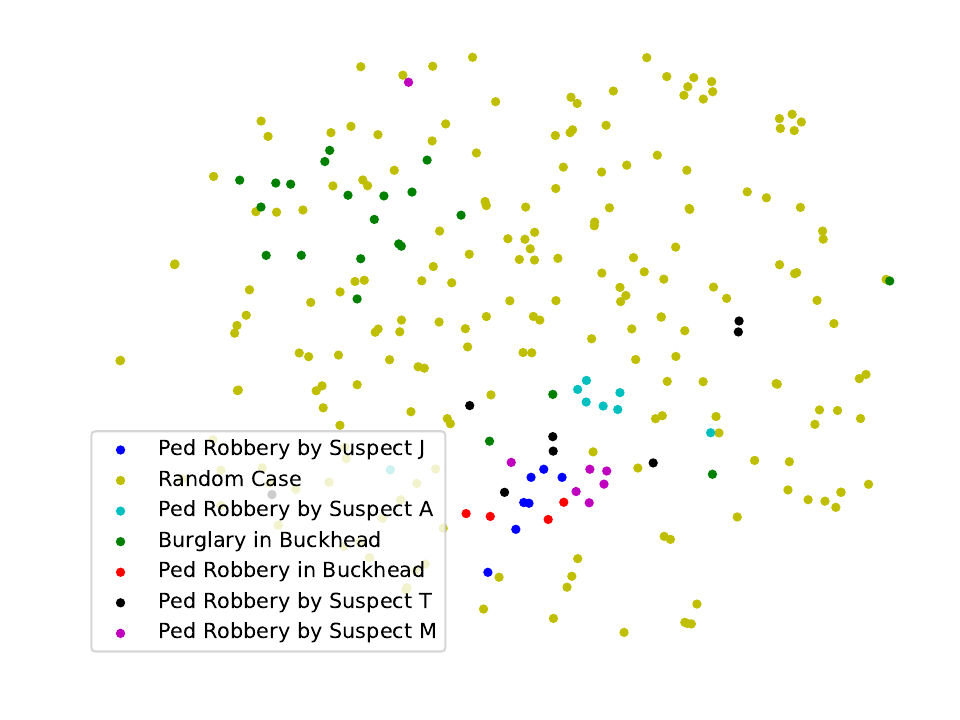}
\caption{vanilla RBM ($\delta = 0$)}
\end{subfigure}
\begin{subfigure}[h]{0.48\linewidth}
\includegraphics[width=\linewidth]{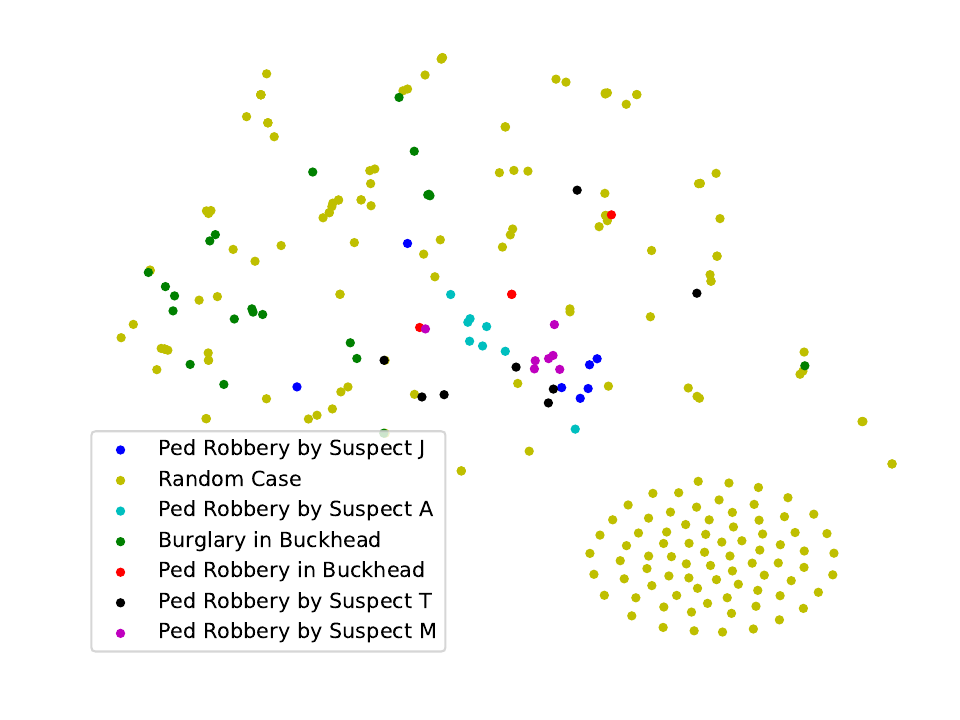}
\caption{regularized RBM with $\delta = 10^{-2}$}
\end{subfigure}
\caption{Comparison of embeddings generated by vanilla RBM and our proposed regularized RBM. Each dot represents a crime incident's embedding, projected on a two-dimensional space via t-SNE [\cite{vanDerMaaten2008}]; we show 56 labeled and 500 random crime incidents in both figures. 
}
\label{fig:tsne}
\end{figure}

The selected keywords (with \textit{TF-IDF} value $> 10^{-2}$) can be observed in the \emph{reconstructed corpus}, which is generated by performing Gibbs sampling according to the conditional probability \eqref{eq:cond-prob-visible} given the text embeddings. Note that the selected subset of keywords in the reconstructed corpus is small, 280 out of 7038 keywords, as desired. 
The selected keywords are supposed to be the most important for embedding. 
Thus, we examine the selected keywords in Figure~\ref{fig:corpus}, together with high \textit{TF-IDF} keywords. 
These keywords correspond to common descriptors in crime reports. The combination of keywords with high TF-IDF values may portray a certain aspect of the incident and hence, help to identify crime linkages. For example, a combination of high TF-IDF keywords, including \emph{home}, \emph{door}, \emph{window}, \emph{stolen}, may be associated with a burglary incident; whereas a combination of \emph{midnight}, \emph{Toyota Corolla}, \emph{pistol} may be associated with an armed robbery incident. 
Some keywords are strong indicators of crime type, for example, \emph{marijuana}, \emph{vandalism}, whereas some keywords may reveal the stage of the investigation of the related crime series; for example, \emph{wasn't sure}, \emph{suspect}, \emph{identified}, \emph{arrestee}, \emph{police custody}, \emph{Miranda}.
In Figure~\ref{fig:tsne}, we also show that 
the embeddings generated by the regularized RBM can better cluster incidents for the identified crime series.
This confirms that our proposed regularized RBM is more effective in clustering police reports than the vanilla RBM by removing irrelevant keywords.

\subsection{Evaluation metrics and procedure} 

We adopt standard performance metrics, including precision, recall, and $F_1$ score. This choice is because linkage detection can be viewed as a binary classification problem, where we aim to identify if there is a linkage between two arbitrary crime incidents in the data. 
Define the set of all truly related incident pairs as $U$, the set of positive incident pairs retrieved by our method as $V$.
Then precision $P$ and recall $R$ are defined as: 
\[
    P = |U \cap V|/|V|,~R = |U \cap V|/|U|,
\] 
where $|\cdot|$ is the number of elements in the set. 
The $F_1$ score combines the \textit{precision} and \textit{recall}: $F_1 = 2 P R / (P + R)$ and the higher $F_1$ score the better. Since numbers of positive and negative pairs in real data are highly unbalanced, we do not use the ROC curve (true positive rate versus false-positive rate) in our setting. 

The evaluation procedure is as follows. The six identified crime series include 56 incidents with more than $1,000$ detected crime linkages between them, which can be used as ``true labels'' for training and validation purposes. We first shuffle the labeled data set randomly and split the crime series into $k$ groups ($k=5$), where each group has at least one identified crime series. For each unique group, we take the group as a hold-out or test data set and take the remaining groups as a training data set. 
Then we fit a model using all the unlabeled data to find the optimal $A$ and optimal $\beta$ is chosen for each location by a grid search in $[10^{-8}, 10^{0}]$ using the labeled training data. 
Lastly, we evaluate the fitted model with optimal $A, \mu, \beta$ on all the labeled test sets and obtain the average $F_1$ score.
Given all possible incident pairs in a group of crime incidents, we retrieve the top $N$ pairs with the highest linkage likelihood $p_{ij}$, $\forall i, j$ values returned our algorithm. If two crime incidents of a retrieved pair were indeed in the same crime series, then it is a success. Otherwise, the pair is un-linked, and it is a misdetection. 
In our data, burglary has 55,278 pairs in total, and 97 are linked, robbery has 60,726 pairs in total, and 231 are linked, mixed has 46,360 pairs in total, 328 of them are linked. 

\subsection{Choice of temporal coefficient $\beta$ by cross-validation}
\label{sec:temporal_coefficient}

\begin{figure}[!t]
\vspace{-.1in}
\centering
\includegraphics[width=.4\linewidth]{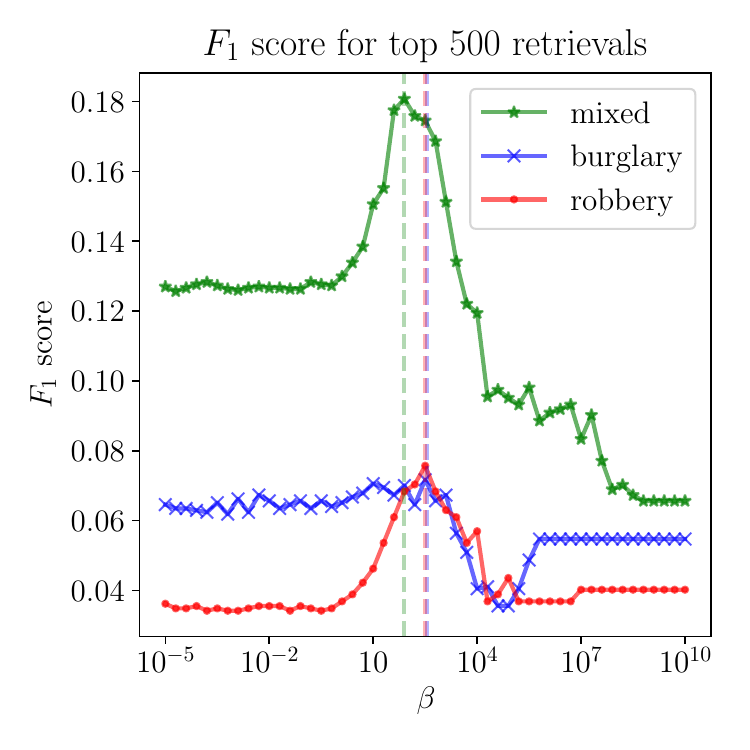}
\caption{Choice of the optimal $\beta$ by $5$-fold cross-validation: $F_1$ scores over the temporal correlation parameter $\beta$ when retrieving the top $N=500$ incident pairs with the highest correlated probabilities $p_{ij}$; results are obtained by repeating over 50 random experiments. }
\label{fig:avg-performance-beta-alpha}
\vspace{-.1in}
\end{figure}

We show that an optimal choice of the parameter $ \beta $ for crime linkage detection exists, which achieves the best bias-variance tradeoff. 
Specifically, we perform the $k$-fold ($k=5$) cross-validation with a grid search to find the optimal $\beta$ using the labeled data.
The tradeoff can be intuitively explained as, in \eqref{eq:estep-2}, when $ \beta $ are too small, the long-range temporal dependence is not captured, and when $ \beta $ are too large, the process may not forget the history. Moreover, in our model, the influence kernel jointly captures spatio-temporal and text influence. When $ \beta $ is too large, the temporal influence may dominate the contribution of textual correlation. An appropriately set temporal coefficient $\beta$ can improve the performance of our method. A real-data example is shown in Figure~\ref{fig:avg-performance-beta-alpha}, where the vertical dash lines in the figures indicate where the model attains its best performance regarding the $F_1$ score (defined in Section \ref{sec:baselines}). We test \texttt{STTPP+RegRBM} using $N=500$ pairs of arbitrarily retrieved results (including both linked and un-linked cases). We note that in the experiments, $\beta \approx 10^2$ leads to the best performance. In practice, when there is a handful of labeled data indicating crime series (like the dataset we have here), we can use this small amount of training data to pre-select an optimal $\beta$ used in our model. 

\subsection{Estimated spatial coefficients $\alpha_{u, v}$ and interpretations}
\label{sec:spatial_coefficient}

\begin{figure}[!t]
\centering
\begin{subfigure}[h]{0.35\linewidth}
\includegraphics[width=\linewidth]{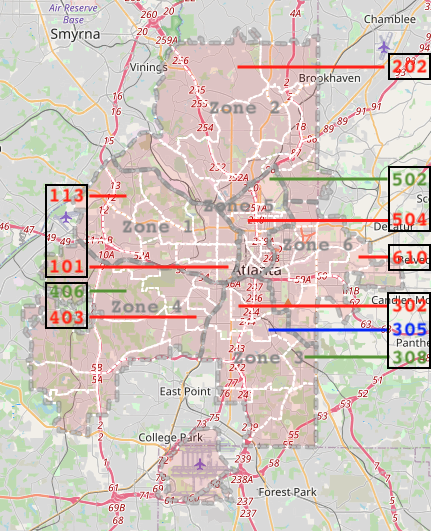}
\caption{Beat map}
\label{fig:beat-map}
\end{subfigure}
\hspace{.25in}
\begin{subfigure}[h]{0.45\linewidth}
\includegraphics[width=\linewidth]{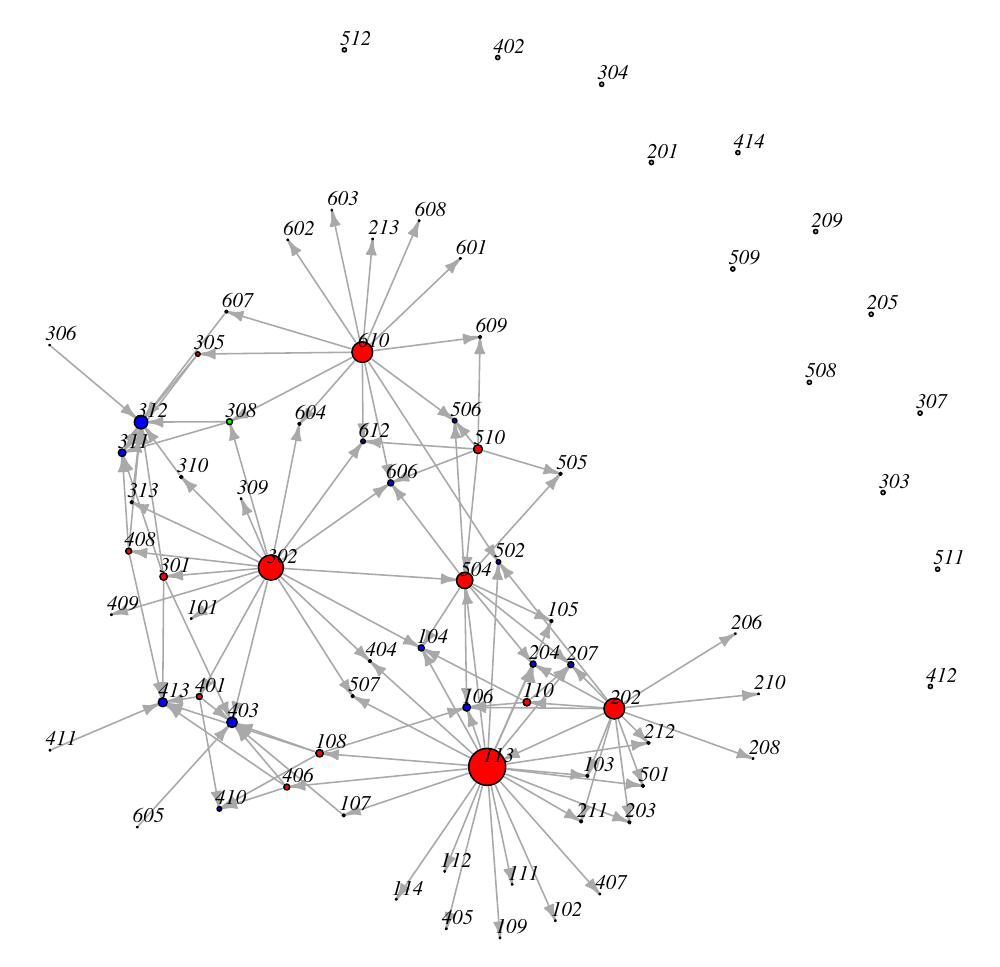}
\caption{Burglary}
\label{fig:graph-burglary}
\end{subfigure}
\vfill
\begin{subfigure}[h]{0.45\linewidth}
\includegraphics[width=\linewidth]{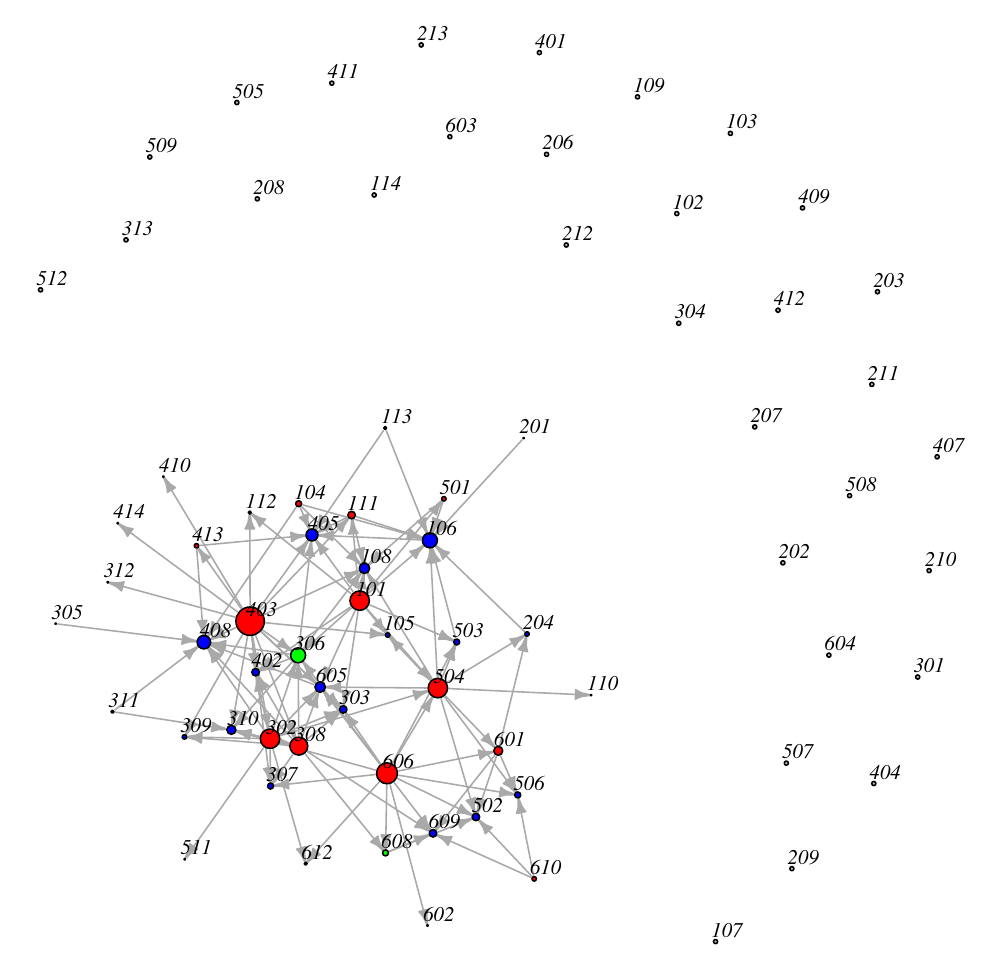}
\caption{Robbery}
\label{fig:graph-robbery}
\end{subfigure}
\begin{subfigure}[h]{0.45\linewidth}
\includegraphics[width=\linewidth]{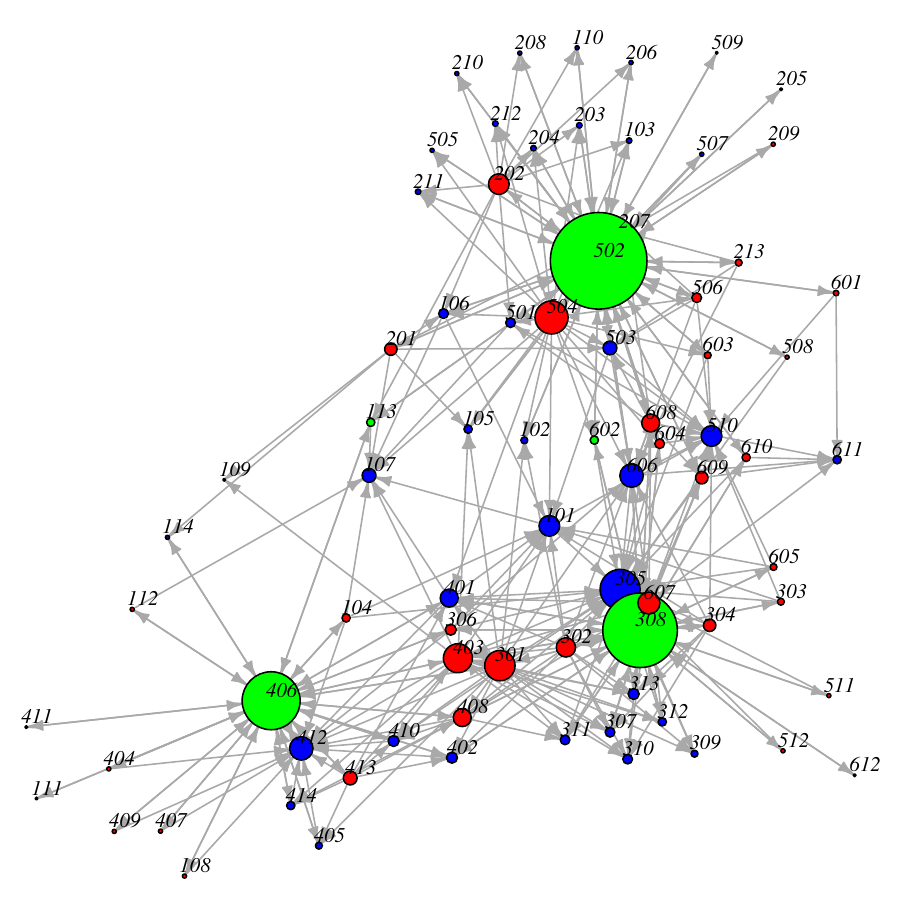}
\caption{Mixed}
\label{fig:graph-mixed}
\end{subfigure}
\caption{(a) Police beat map of Atlanta: regions with white boundary are beats; the ID of beats with significant influence learned by our model has been highlighted. (b-d) Directed graphs corresponding to the estimated $\widehat A$ for three scenarios, where each node represents a beat and each edge corresponds to $\alpha_{u,v} > 0.5$. The red nodes are the beats with larger outdegree than indegree, indicating that the incidents in these beats are more likely to trigger subsequent incidents in the connected beats; the blue nodes are the opposite; the green nodes indicate the beats have equal indegree and outdegree, and the grey nodes are isolated. The dots' size indicates the indegree and outdegree differences. We notice that certain beats are more influential than others.
}
\label{fig:a-graph}
\end{figure}

The EM algorithm can estimate the spatial coefficient matrix $A$, capturing the directional influence between locations. The magnitude of the coefficient captures how large the influence is. In our experiments, we randomly initialize the coefficients in $(0, 1)$. 

Now we visualize the estimated $A$ in Figure~\ref{fig:a-graph}, treating it as the adjacency matrix of a weighted and directed graph, where nodes represent beats and edges represent the spatial influences. We threshold the estimated spatial coefficients and only keep edges if $a_{ij} > 0.5$. For burglary and robbery (Figure~\ref{fig:graph-burglary}, \ref{fig:graph-robbery}), some beats are isolated and have no connection with any other beats. A few beats, indicated by large red dots in the graph, have a dominating influence on their surrounding beats. 
An interesting observation is that 
the linked incidents in the same crime series usually occurred within a few (2 to 4) neighboring beats, indicating that the perpetrators usually confine their criminal activities in a relatively small area of the city, which they might be familiar with.
The situation becomes more complicated when we consider all types of cases (more than 160 categories of crimes) altogether, as shown in Figure~\ref{fig:graph-mixed}. 
Also, some of the beats, such as 113, 202, 302, and 610, are the crime hotspots, which are very likely to trigger subsequent crime incidents. Increasing patrols in these regions may help to curb future crime and enhance the safety of the city.

\begin{figure}[!t]
\centering
\begin{subfigure}[h]{.325\linewidth}
    \includegraphics[width=\linewidth]{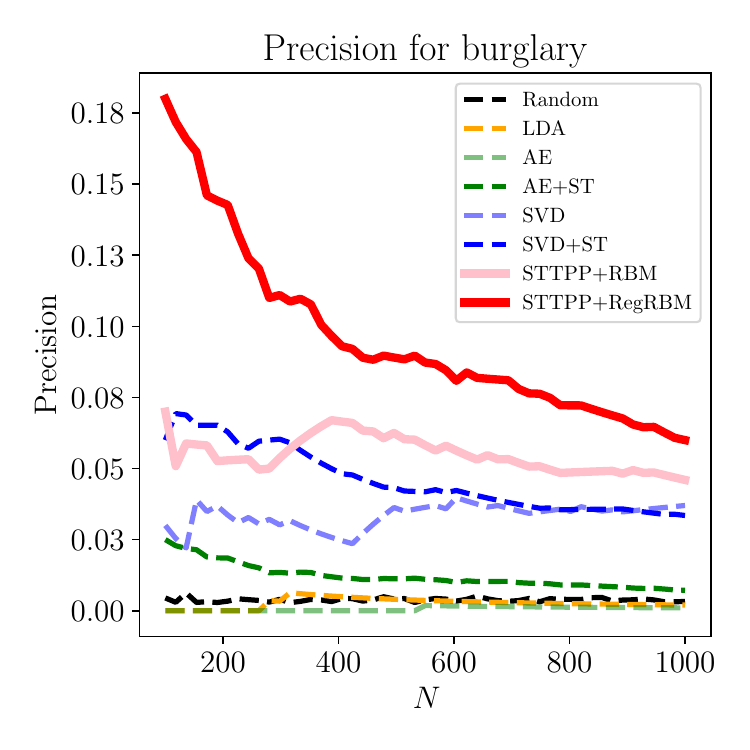}
\end{subfigure}
\begin{subfigure}[h]{.325\linewidth}
    \includegraphics[width=\linewidth]{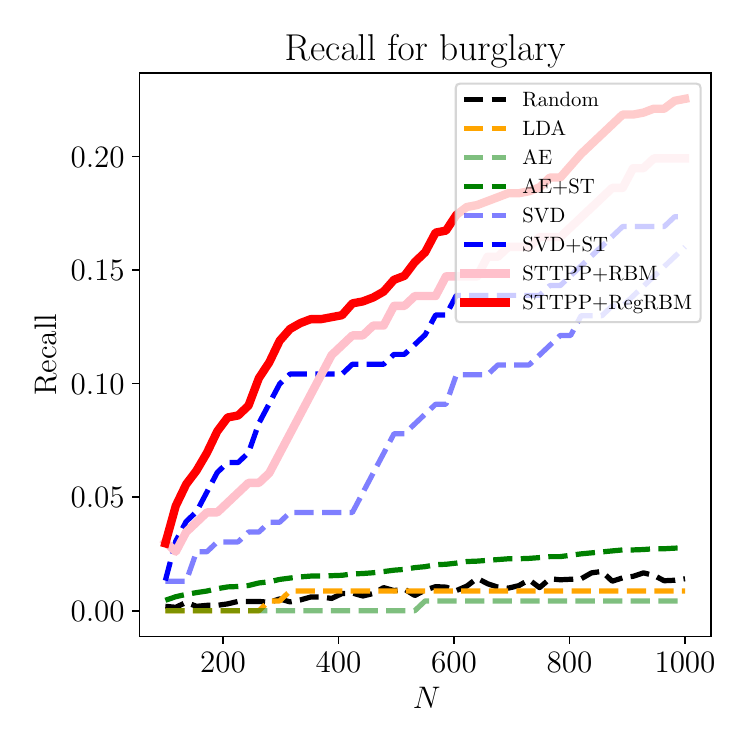}
\end{subfigure}
\begin{subfigure}[h]{.325\linewidth}
    \includegraphics[width=\linewidth]{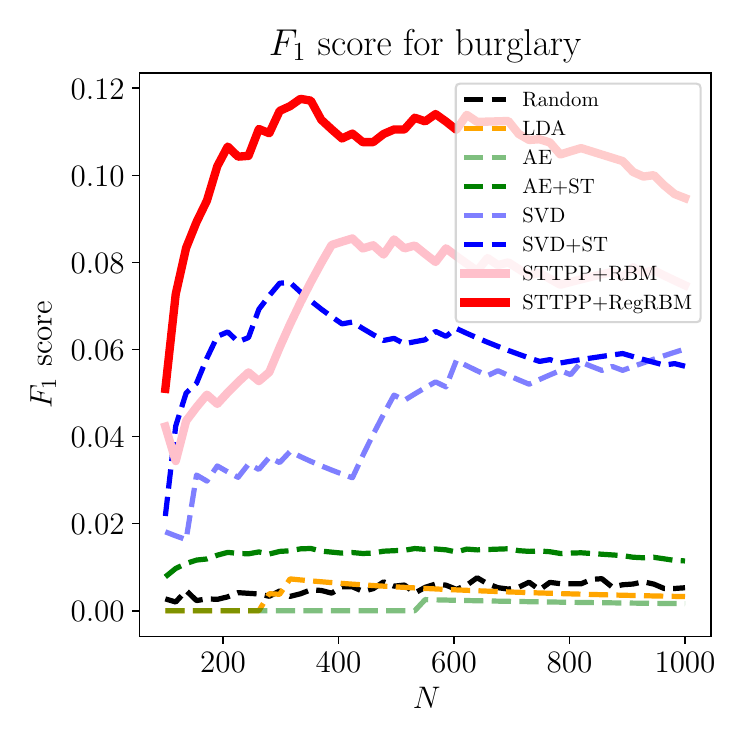}
\end{subfigure}
    \vfill
\begin{subfigure}[h]{.325\linewidth}
    \includegraphics[width=\linewidth]{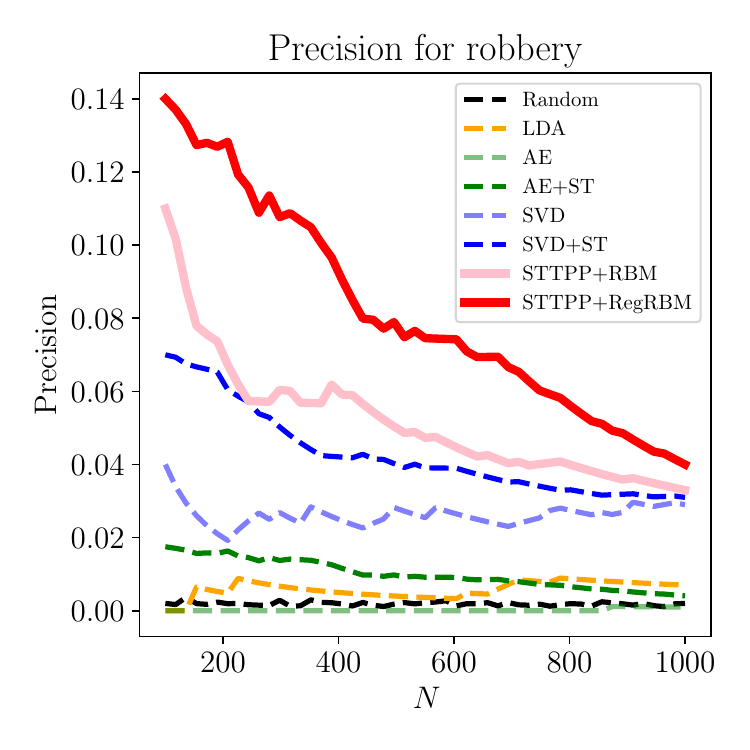}
\end{subfigure}
\begin{subfigure}[h]{.325\linewidth}
    \includegraphics[width=\linewidth]{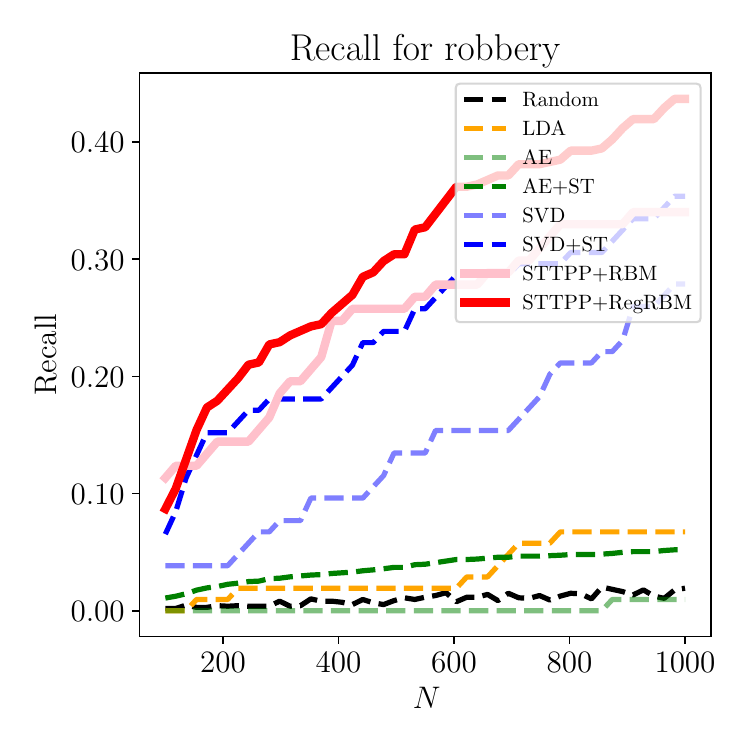}
\end{subfigure}
\begin{subfigure}[h]{.325\linewidth}
    \includegraphics[width=\linewidth]{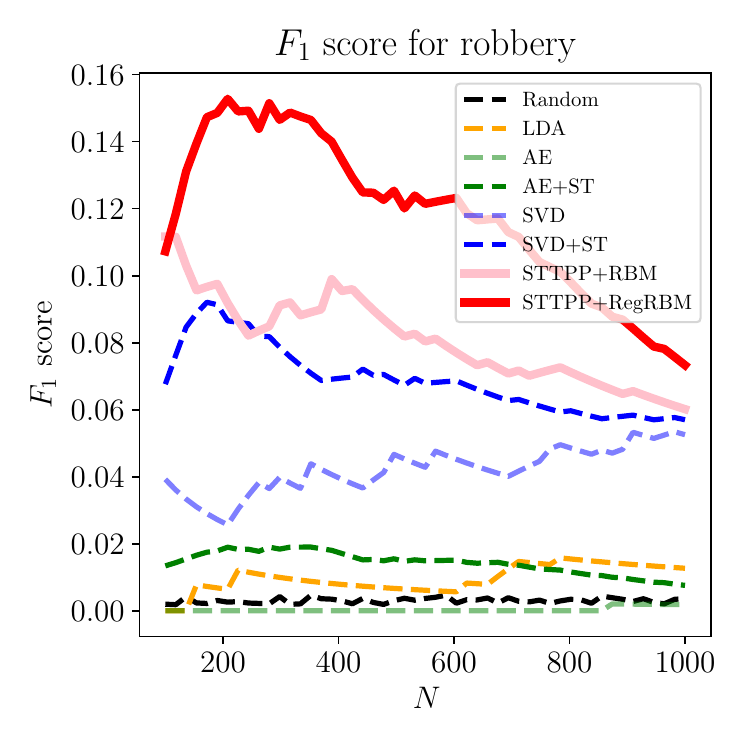}
\end{subfigure}
    \vfill
\begin{subfigure}[h]{.325\linewidth}
    \includegraphics[width=\linewidth]{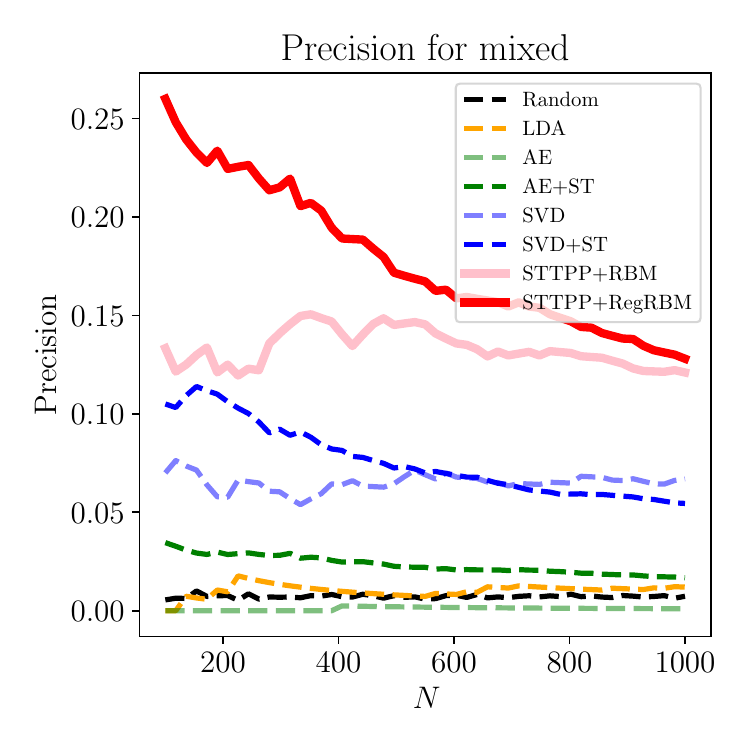}
\end{subfigure}
\begin{subfigure}[h]{.325\linewidth}
    \includegraphics[width=\linewidth]{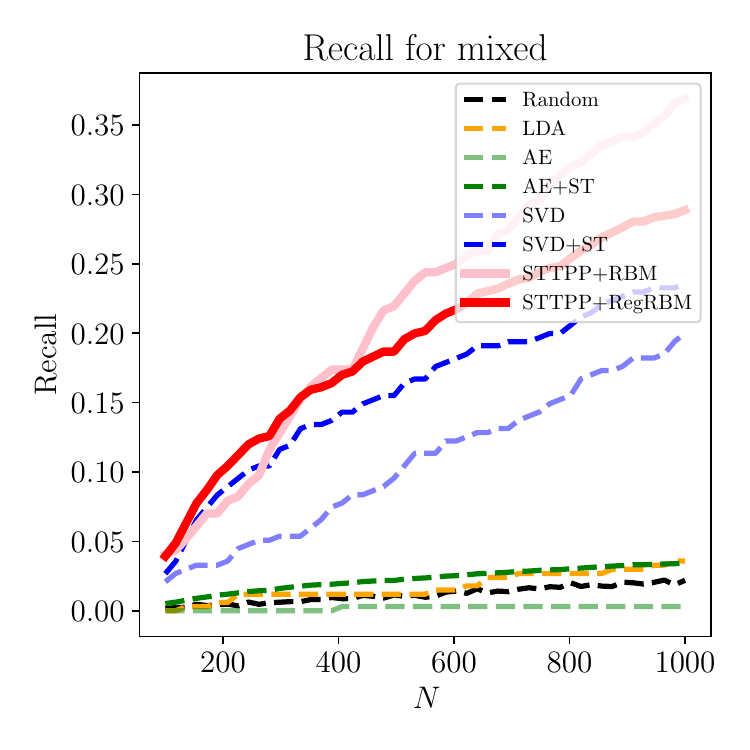}
\end{subfigure}
\begin{subfigure}[h]{.325\linewidth}
    \includegraphics[width=\linewidth]{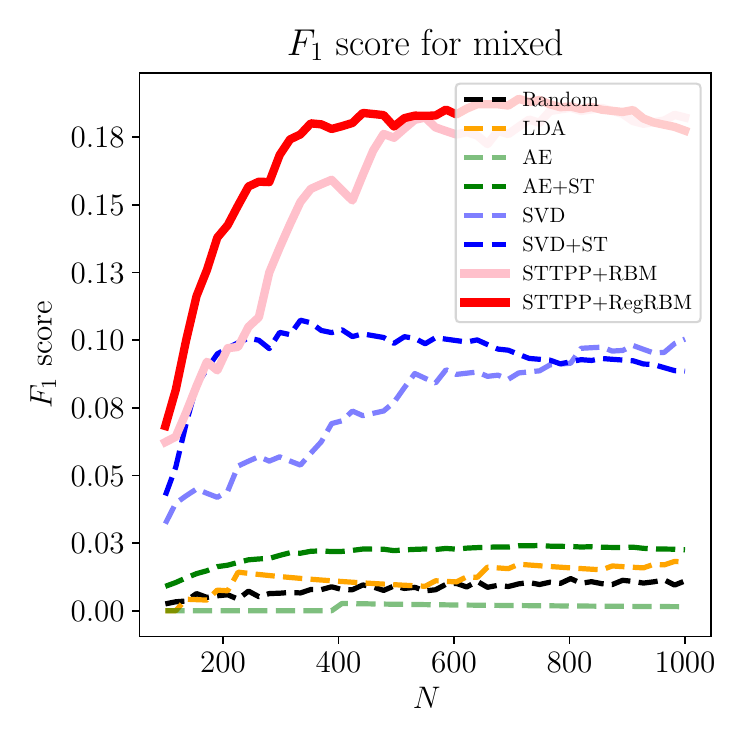}
\end{subfigure}
\caption{Comparison between our method \texttt{STTPP+RegRBM} and baselines with respect to precision, recall and $F_1$ scores for crime linkage detection. We consider a different number of retrievals on data group {\it burglary} (the first row), {\it robbery} (the second row), and {\it mixed} (the third row). The vertical dash lines indicate the location of their best performance.}
\label{fig:baselines}
\end{figure}

\subsection{Comparison with alternative methods}
\label{sec:baselines}

We compare our method, referred to as the  \texttt{STTPP+RegRBM}, to the vanilla RBM without regularization referred to as the (\texttt{STTPP+RBM}), as well as other alternative methods. We may consider  crime linkage detection as an information retrieval task: given a new incident, we would like to find a number of the most relevant incidents from historical data.
We compare with related alternative methods, including Latent Semantic Analysis (performed by Singular Value Decomposition) and  Latent Dirichlet Allocation (\texttt{LDA}) (commonly used in natural language processing). We also compare with Autoencoder (\texttt{AE}), a neural network-based embedding technique.
However, these alternative methods learn embeddings for documents in feature space without considering spatio-temporal information. 
Hence, we extend \texttt{AE} and \texttt{SVD} by including the spatio-temporal information as part of their inputs, which are concatenations of the Bag-of-Words feature vector and incident time and location (latitude and longitude); each dimension of the input is normalized to $[0, 1]$. We refer to these two methods as \texttt{AE+ST} and \texttt{SVD+ST}, respectively.
As a sanity check, we also consider the \texttt{random}-pick strategy as one of the baselines, which randomly selects the subset of variables to be incorporated in the model. To compare with embedding methods, we compute their inner products in the embedding space for each pair of incidents as a similarity score. Based on this, we find the most similar pairs as the retrieval results.


As shown in Figure~\ref{fig:baselines}, our \texttt{STTPP} methods achieves a much higher $F_1$ score than other methods. 
The proposed method attains its best performance at $N=314$, $N=207$, and $N=389$ on burglary, robbery, and mixed datasets. 
This indicates that properly incorporating spatio-temporal information will drastically improve the accuracy of linkage detection. 
In particular, our proposed \texttt{STTPP+RegRBM} greatly outperforms \texttt{STTPP+RBM} (without keyword selection) on single-category crimes, including robbery and burglary. This may be because the \emph{M.O.}s are distributed around a small set of keywords in these cases, and keywords selection plays a critical role in the feature extraction.

\section{Discussions}
\label{sec:conclusion}

This paper introduced a new framework for modeling police incidents as \textit{spatio-temporal-textual} incidents and demonstrated its usage for crime linkage detection. We addressed the challenge of a lack of labeled related incidents (and thus, we face an unsupervised learning problem). We developed a model based on the multivariate marked Hawkes processes, with marks being the incident textual descriptions' embeddings. We incorporated the textual similarity as a component in the point process intensity function while still enjoy computationally tractable likelihood function. We also developed a new embedding technique with keywords selection for text modeling using a regularized Restricted Boltzmann Machine (RBM). The proposed method was validated using a real dataset and compared with alternative methods. We want to remark that the proposed algorithm has been adopted by the Atlanta Police Department and implemented in their AWARE system in 2018. Although we focus on police report analysis in this paper, our model can be used for other types of \textit{spatio-temporal-textual} incident data, such as social media data and electronic health records.

As the linked crime series are hand-labeled by crime analysts, there are likely cases linking to the existing crime series; however, they are not identified yet. We could not consider these in the performance evaluation (due to the lack of the ground truth). It could happen that the algorithm actually detects linkages that are unknown but indeed exist. Nevertheless, the algorithm is shown to be effective based on known labeled cases and provides a potentially helpful tool for crime analysts to cast their attention to unknown but potential links.

We believe our approach based on embedding enjoys certain robustness. The text of the police reports is ``noisy'', in a highly unstructured form with variable length, and are entered by different police officers with personal styles. The embedding is based on a transformation of the raw text of a police report into a fixed-length Bag-of-Words vector and then mapped using the regularized RBM, which captures the intrinsic similarities between text documents rather than apparent keywords. The other aspects of robustness for crime data analysis, such as missing data, are indeed very interesting and can be a topic of further research.

Finally, we want to remark that the crime linkage found by performing analysis based on the Hawkes processes model can be treated as a form of Granger causality [\cite{pmlr-v48-xuc16}], which is a weaker causal inference for time series and does not consider confounding factors. For instance, reports from the same officer may be similar and reflected the officer’s opinion. We believe that using embedding rather than the raw text bag-of-words features will alleviate such an issue to a certain extent since it captures the underlying similarities of crime incidents, as demonstrated in our real-data examples. Further causal inference analysis for crime linkage detection can be a future research direction. Another future direction is to extend the pairwise evaluation for crime linkage detection to consider higher-order interactions among incidents.

\bibliographystyle{plainnat}
\bibliography{refs}

\newpage
\appendix

 \section{Deriving the log-likelihood function of spatio-temporal-textual point processes}
 \label{append:sttpp-likelihood}

 Now we derive the likelihood function of the model. Suppose that there are $n$ incidents before the time horizon $T$. For the notational simplicity, let $z_i = (t_i, k_i, \boldsymbol{h}_i)$. Let 
 \[F^*(t) = \mathbb{P}\{t_{n+1} < t, k_{n+1}, \boldsymbol{h}_{n+1} | \mathcal{H}_t\}\] be the conditional probability that next incident $t_{n+1}$ happens before $t$ given the filtration generated by the historical incidents $\mathcal{H}_t$ and let $f^*(t)$ be the corresponding conditional probability density function. The conditional intensity function [\cite{BaBoMa2013}] for a given $k$ and $\boldsymbol{h}$ is defined by
 \[
 \lambda_t = f^*(t) / (1 - F^*(t)).
 \]
 We can show
 $
 \lambda_t 
 = - d \log (1 - F^*(t))/dt.
 $
 Hence, \[\int_{t_n}^{t} \lambda_{\tau} d\tau = - \log (1- F^*(t)),\] where $F^*(t) = 0$ for $t_n \le t < t_{n+1}$, since the $(n+1)$th incident does not exist at time $t_n$. Therefore, 
 \[
 F^*(t) = 1 - \exp\left\{ - \int_{t_n}^{t} \lambda_{\tau} d\tau \right\},\] and 
 \[f^*(t) = \lambda_t \exp \left\{ - \int_{t_n}^{t} \lambda_\tau d\tau \right\}.
 \]
 Now we write the joint distribution of a sequence by decomposing into the product of a series of conditional distributions: 
 \begin{equation}
 \begin{aligned}
  \mathbb{P}\{Z_1 = z_1, \dots, Z_n = z_n\}= &~ (1-F^*(T)) \prod_{i=1}^{n} f(t_i|\mathcal{H}_{t_i}) = (1-F^*(T)) \prod_{i=1}^{n} f^*(t_i) \\
 = &~ \left ( \prod_{i=1}^{n} \lambda_{t_i}^{s_i}(\boldsymbol{h}_i) \right ) \exp\left\{- \sum_{k=1}^{d} \sum_{\boldsymbol{h} \in \Omega} \int_{0}^{T} \lambda_{\tau}^k(\boldsymbol{h}) d\tau\right\},
 \end{aligned}
 \label{eq:pp_likelihood}
 \end{equation}
 where $\Omega \subseteq \{0, 1\}^m$ is the space of embedding (the size of the space is usually relatively small $|\Omega| \ll 2^m$).

 From the \eqref{eq:pp_likelihood}, we can derive the log-likelihood function of spatio-temporal-textual point processes as follows:
 \begin{equation}
     \ell(A) = \log \left ( \prod_{i=1}^{n} \lambda_{t_i}^{s_i}(\boldsymbol{h}_i) \right ) - \sum_{k=1}^{d} \sum_{\boldsymbol{h} \in \Omega} \int_{0}^{T} \lambda_{\tau}^k(\boldsymbol{h}) d\tau,
     \label{eq:derivation-log-likelihood-1}
 \end{equation}
 where the first term on the right side of \eqref{eq:derivation-log-likelihood-1} is obtained by
 \begin{align*}
     \log \left ( \prod_{i=1}^{n} \lambda_{t_i}^{s_i}(\boldsymbol{h}_i) \right )
     = \sum_{i=1}^n \log \left( \mu_{s_i} + \sum_{j=1}^{i-1} \alpha_{s_i,s_j} \beta e^{-\beta (t_i - t_j)} \tilde{\boldsymbol{h}}_i^\intercal \tilde{\boldsymbol{h}}_j \right).
 \end{align*}
 The second term on the right side of \eqref{eq:derivation-log-likelihood-1} is obtained by
 \begin{align*}
     & \sum_{k=1}^{d} \sum_{\boldsymbol{h} \in \Omega} \int_{0}^{T} \lambda_{\tau}^k(\boldsymbol{h}) d\tau\\
     =& \sum_{k=1}^{d} \sum_{\boldsymbol{h} \in \Omega} \int_{0}^{T} \left(\mu_k + \sum_{j: t_j < \tau} \alpha_{k,s_j} \beta e^{-\beta (\tau - t_j)} \tilde{\boldsymbol{h}}^\intercal \tilde{\boldsymbol{h}}_j\right) d\tau\\
     =& \sum_{k=1}^d \mu_k |\Omega| T + \sum_{k=1}^{d} \sum_{\boldsymbol{h} \in \Omega} \int_{0}^{T} \left(\sum_{j: t_j < \tau} \alpha_{k,s_j} \beta e^{-\beta (\tau - t_j)} \tilde{\boldsymbol{h}}^\intercal \tilde{\boldsymbol{h}}_j\right) d\tau\\
     =& \sum_{k=1}^d \mu_k |\Omega| T + \sum_{k=1}^{d} \sum_{\boldsymbol{h} \in \Omega} \int_0^T \left( \sum_{j=1}^n \mathbb{I}\{\tau > t_j\} \alpha_{k,s_j} \beta e^{-\beta (\tau-t_j)} \tilde{\boldsymbol{h}}^\intercal \tilde{\boldsymbol{h}}_j \right) d\tau\\
     =& \sum_{k=1}^d \mu_k |\Omega| T + \sum_{k=1}^{d} \sum_{\boldsymbol{h} \in \Omega} \sum_{j=1}^n \int_{t_j}^T \left( \alpha_{k,s_j} \beta e^{-\beta (\tau-t_j)} \tilde{\boldsymbol{h}}^\intercal \tilde{\boldsymbol{h}}_j \right) d\tau \\
     =& \sum_{k=1}^d \mu_k |\Omega| T + \sum_{k=1}^{d} \sum_{\boldsymbol{h} \in \Omega} \sum_{j=1}^n \alpha_{k,s_j} \beta \left( \int_{t_j}^T e^{-\beta (\tau-t_j)} d\tau \right) \tilde{\boldsymbol{h}}^\intercal \tilde{\boldsymbol{h}}_j \\
     =& \sum_{k=1}^d \mu_k |\Omega| T + \sum_{k=1}^{d} \sum_{\boldsymbol{h} \in \Omega} \sum_{j=1}^n \alpha_{k,s_j} \left( 1 - e^{-\beta (T - t_j)} \right) \tilde{\boldsymbol{h}}^\intercal \tilde{\boldsymbol{h}}_j.
 \end{align*}
 Therefore, we can obtain the log-likelihood function as in \eqref{eq:pp_log_likelihood}. 

 \section{The lower bound of the log-likelihood function}
 \label{append:lower-bound-log-likehood}

 Given auxiliary variables $\{p_{ij}\}$, which satisfy $\forall i, \sum_{j=1}^i p_{ij} = 1, p_{ij} > 0$, we are able to derive the lower bound of the first term on the right side of \eqref{eq:pp_log_likelihood}:
 \begin{align*}
     & \sum_{i=1}^n \log \left( \mu_{s_i} + \sum_{j=1}^{i-1} \alpha_{s_i,s_j} \beta e^{-\beta (t_i - t_j)} \tilde{\boldsymbol{h}}_i^\intercal \tilde{\boldsymbol{h}}_j \right)\\
     = & \sum_{i=1}^n \log \left( p_{ii} \frac{\mu_{s_i}}{p_{ii}} + \sum_{j=1}^{i-1} p_{ij} \frac{\alpha_{s_i,s_j} \beta e^{-\beta (t_i - t_j)} \tilde{\boldsymbol{h}}_i^\intercal \tilde{\boldsymbol{h}}_j}{p_{ij}} \right)\\
     \ge & \sum_{i=1}^n \Biggl( p_{ii}\log(\mu_{s_i}) + \sum_{j=1}^{i-1} p_{ij} \log \left( \alpha_{s_i, s_j} \beta e^{-\beta (t_i - t_j)} \tilde{\boldsymbol{h}}_i^\intercal \tilde{\boldsymbol{h}}_j \right) - p_{ii} \log p_{ii} - \sum_{j=1}^{i-1} p_{ij} \log p_{ij} \Biggr).
 \end{align*}
 The last inequality is due to Jensen's inequality.

 \section{Derive EM algorithm}
 \label{append:grads-em}

 We denote the lower bound of the log-likelihood function as $\ell'(A)$, i.e.,
 \begin{equation}
 \begin{aligned}
     \ell'(A) = & \sum_{i=1}^n \Biggl( p_{ii}\log(\mu_{s_i}) + \sum_{j=1}^{i-1} p_{ij} \log \left( \alpha_{s_i, s_j} \beta e^{-\beta (t_i - t_j)} \tilde{\boldsymbol{h}}_i^\intercal \tilde{\boldsymbol{h}}_j \right) \\
     - & \sum_{j=1}^i p_{ij} \log p_{ij} \Biggr) - \sum_{k=1}^d \mu_k |\Omega| T - \sum_{k=1}^{d} \sum_{j=1}^n \sum_{\boldsymbol{h} \in \Omega} \alpha_{k,s_j} \left( 1 - e^{-\beta (T - t_j)} \right) \tilde{\boldsymbol{h}}^\intercal \tilde{\boldsymbol{h}}_j.
     \label{eq:tmp_log_likelihood_lowerb}
 \end{aligned}
 \end{equation}
 First we derive the optimal $\hat{\alpha}_{u,v}$ by setting the partial derivative of $\ell'$ with respect to $\alpha_{u,v}$ to be 0:
 \begin{align*}
     \frac{\partial \ell'}{\partial \alpha_{u,v}}
     = & \frac{1}{\alpha_{u,v}} \left(\sum_{i=1}^{n} \sum_{j=1}^{i-1} \mathbb{I}\{s_i = u, s_j = v\} p_{ij} \right) - \sum_{j=1}^{n} \mathbb{I}\{s_j = v\} (1 - e^{-\beta (T - t_j)}) \sum_{\boldsymbol{h} \in \Omega} \tilde{\boldsymbol{h}}^\intercal \tilde{\boldsymbol{h}}_j = 0.
 \end{align*}
 Solving the equation, we obtain the optimal $\hat{\alpha}_{u,v}$:
 \begin{align*}
     \hat{\alpha}_{u,v} = \frac{\sum_{i=1}^{n} \sum_{j=1}^{i-1} \mathbb{I}\{s_i = u, s_j = v\} p_{ij} }{\sum_{j=1}^{n} \mathbb{I}\{s_j = v\} (1 - e^{-\beta (T - t_j)}) \sum_{\boldsymbol{h} \in \Omega} \tilde{\boldsymbol{h}}^\intercal \tilde{\boldsymbol{h}}_j},~\forall u, v = 1, \dots, d.
 \end{align*}

 Next, we derive the optimal $\{p_{ij}\}_{j\le i}$. Note that 
 \begin{equation}
 \forall i,\ p_{ii} = 1 - \sum_{j=1}^{i-1} p_{ij}.
 \label{eq:tmp_constraint}
 \end{equation}
 Substitute \eqref{eq:tmp_constraint} into \eqref{eq:tmp_log_likelihood_lowerb}, we have
 \begin{align*}
     \ell'(A) = & \sum_{i=1}^n \Biggl( (1 - \sum_{j=1}^{i-1} p_{ij}) \log(\mu_{s_i}) + \sum_{j=1}^{i-1} p_{ij} \log \left( \alpha_{s_i, s_j} \beta e^{-\beta (t_i - t_j)} \tilde{\boldsymbol{h}}_i^\intercal \tilde{\boldsymbol{h}}_j \right) - \sum_{j=1}^{i-1} p_{ij} \log p_{ij} \\ 
     - & (1 - \sum_{j=1}^{i-1} p_{ij}) \log (1 - \sum_{j=1}^{i-1} p_{ij})\Biggr) - \sum_{k=1}^d \mu_k |\Omega| T - \sum_{k=1}^{d} \sum_{j=1}^n \sum_{\boldsymbol{h} \in \Omega} \alpha_{k,s_j} \left( 1 - e^{-\beta (T - t_j)} \right) \tilde{\boldsymbol{h}}^\intercal \tilde{\boldsymbol{h}}_j.
 \end{align*}
 Set the partial derivative of $\ell'$ with respect to $p_{ij}$ equal to 0 for $j < i, i = 1, \dots, n$:
 \begin{align*}
     \frac{\partial \ell'}{\partial p_{ij}}
     = & - \log(\mu_{s_i}) + \log \left( \alpha_{s_i, s_j} \beta e^{-\beta (t_i - t_j)} \tilde{\boldsymbol{h}}_i^\intercal \tilde{\boldsymbol{h}}_j \right) - \log p_{ij} + \log(1 - \sum_{l=1}^{i-1} p_{il}) = 0.
 \end{align*}
 Let $h_{ij} = \alpha_{s_i, s_j} \beta e^{-\beta (t_i - t_j)} \tilde{\boldsymbol{h}}_i^\intercal \tilde{\boldsymbol{h}}_j$. We have
 \[
     \frac{p_{ij}}{1 - \sum_{j=1}^{i-1} p_{ij}} = \frac{h_{ij}}{\mu_{s_i}}
     \Rightarrow \mu_{s_i} p_{ij} = h_{ij} \left(1 - \sum_{l = 1}^{i-1} p_{il}\right).
 \]
 Add $- h_{ij} p_{ii}$ to both sides. Due to constraint \eqref{eq:tmp_constraint}, the above equation can written as  
 \[
     \mu_{s_i} p_{ij} - h_{ij} p_{ii} = 0.
 \]
 Thus,
 \begin{equation}
     p_{ij} = \frac{h_{ij}}{\mu_{s_i}} p_{ii}.
     \label{eq:pij-use-pii}
 \end{equation}
 Now, sum over all $j \le i - 1$, we obtain
 \[
     \sum_{j=1}^{i-1} p_{ij} = \sum_{j=1}^{i-1} \frac{h_{ij}}{\mu_{s_i}} p_{ii}.
 \]
 Use \eqref{eq:tmp_constraint} again, we obtain 
 \[
     1 - p_{ii} = \sum_{j=1}^{i-1} \frac{h_{ij}}{\mu_{s_i}} p_{ii}.
 \]
 From this we can solve the optimal $\hat{p}_{ii}$ by: 
 \[
     \hat{p}_{ii} = \frac{\mu_{s_i}}{\mu_{s_i} + \sum_{j=1}^{i-1} h_{ij}}, ~\forall i = 1, \dots, n.
 \]
 Finally, use \eqref{eq:pij-use-pii}, we derive the optimal $\hat{p}_{ij}$ given by:
 \[
     \hat{p}_{ij} = \frac{h_{ij}}{\mu_{s_i} + \sum_{j=1}^{i-1} h_{ij}}, ~\forall j < i, i = 1,\dots, n.
 \]

 \section{Gradients of regularized likelihood function}
 \label{append:reg-rbm-loglikelihood}

 From the standard result of the RBM, we can derive the gradient of the log-likelihood function (without the penalty term), given by
 \[
 \left< X_l H_j\right >_{p(\boldsymbol{H}|\boldsymbol{X})q(\boldsymbol{X})}  - \left < X_l H_j\right >_{p(\boldsymbol{X}, \boldsymbol{H})}
 \]
 Now consider the penalty term. Note that
 \[
  \mathbb P(X_l \geq \tau | \boldsymbol{H}; \theta)
  = \int_\tau^\infty \mathcal N(z; b_l + \sigma \sum_{j=1}^m H_j w_{lj}, \sigma^2) dz.
 \]
 Now we derive the gradient of the penalty term. For a given $l$ and any $k = 1, \ldots, m$, we have that
 \[
 \frac{\partial \mathbb P(X_l \geq \tau | \boldsymbol{H}; \theta)}{\partial w_{jk}} = 0, \quad \forall l\neq j.
 \]
 We also have
 \begin{equation}
 \begin{split}
 &\frac{\partial \mathbb P(X_l \geq \tau | \boldsymbol{H}; \theta)}{\partial w_{lk}}\\
 = &~ \frac{1}{\sigma \sqrt{2\pi}}\cdot  \int_\tau^{\infty} \frac{\partial \exp\left\{ -\frac{1}{2\sigma^2} (z- b_l- \sigma \sum_{j=1}^m H_j w_{lj})^2 \right\}}{\partial w_{lk}} dz\\
 = &~  \frac{1}{\sigma \sqrt{2\pi}}\cdot
  \int_\tau^{\infty} \exp\left\{ -\frac{1}{2\sigma^2} (z- b_l- \sigma \sum_{j=1}^m H_j w_{lj})^2 \right\}\cdot \frac{H_k}{\sigma}\cdot(z- b_l- \sigma \sum_{j=1}^m H_j w_{lj}) dz.
 \end{split}
 \end{equation}
 Let $u = (z- b_l- \sigma \sum_{j=1}^m H_j w_{lj})/\sigma$, and $\tau'_l=\tau - b_l- \sigma \sum_{j=1}^m H_j w_{lj}$. Note that $dz = \sigma du$. After change of variable, the above equation becomes
 \begin{equation}
 \begin{split}
 \frac{\partial \mathbb P(X_l \geq \tau | \boldsymbol{H}; \theta)}{\partial w_{lk}}&= H_k  \cdot \frac{1}{\sqrt{2\pi}}
  \int_{\tau'}^{\infty} u \exp\left\{ -\frac{1}{2} u^2 \right\}\cdot
 du\\
 &= H_k \cdot \frac{\phi(\tau'_l)}{1-\Phi(\tau'_l)}. \\
 \end{split}
 \end{equation}
 The calculation above is done by realizing it corresponds to finding the mean of the truncated normal distribution greater than $\tau'$, and using $\mathbb E [X|X>a] = \phi(a)/(1-\Phi(a))$, for $X$ being a standard normal random variable.

 Similarly, for given $l$, we obtain
 \begin{equation}
 \begin{split}
 &\frac{\partial \mathbb P(X_l \geq \tau | \boldsymbol{H}; \theta)}{\partial b_{l}}\\
 = &~ \frac{1}{\sigma \sqrt{2\pi}}  \int_\tau^{\infty} \frac{\partial 
 \exp\left\{ -\frac{1}{2\sigma^2} (z- b_l- \sigma \sum_{j=1}^m H_j w_{lj})^2 \right\} }{\partial b_{l}} dz \\
 = &~  \frac{1}{\sigma \sqrt{2\pi}} 
  \int_\tau^{\infty} \exp\left\{ -\frac{1}{2\sigma^2} (z- b_l- \sigma \sum_{j=1}^m H_j w_{lj})^2 \right\} \frac{1}{\sigma^2} (z- b_l- \sigma \sum_{j=1}^m H_j w_{lj}) dz\\
 = &~  \frac{1}{\sigma^2} \frac{1}{\sqrt{2\pi}} \int_{\tau'}^\infty
 u \exp\left\{ -\frac{u^2}{2}  \right\} \\
 = & ~ \frac{1}{\sigma^2} \frac{\phi(\tau')}{1-\Phi(\tau')}.
 \end{split}
 \end{equation}
 Moreover, \[\frac{\partial \mathbb P(X_l \geq \tau | \boldsymbol{H}; \theta)}{\partial b_{j}}=0, \quad \forall l\neq j\]
 Similarly, for given $k$, we obtain
 \[
 \frac{\partial \mathbb P(X_l \geq \tau | \boldsymbol{H}; \theta)}{\partial c_k} =0, \quad \forall l, k.
 \]
 Putting the above together we obtain the final results.

 \section{Non-parametric background rate estimation}
 \label{append:background}

 Here we also provide additional numerical results in Table~\ref{tab:compare-to-stochastic-declustering} to show that our approach can achieve similar results to the traditional stochastic declustering while being computationally efficient. In particular, for the same training set with 500 crime incidents and given the same $\beta$, our algorithm only takes six iterations of (8a-c) in 23 minutes to reach the convergence, whereas the stochastic declustering with randomly initialized $\mu_k$ takes 51 iterations in around 200 minutes.

 \begin{table}[!h]
 \centering
 \caption{The error between the estimation results using the proposed method and the stochastic declustering.}
 \label{tab:compare-to-stochastic-declustering}
 \resizebox{1.\textwidth}{!}{%
 \begin{threeparttable}
 \begin{tabular}{lcccccc}
 \toprule[1pt]\midrule[0.3pt]
 Parameters & Mean AE & Mean APE & Max AE & Max APE & Min AE & Min APE \\ \hline
 \multicolumn{1}{l}{$\mu_k$} & 0.137 & 4.231\% & 0.524 & 17.210\% & 0.011 & 0.450\% \\
 \multicolumn{1}{l}{$\alpha_{u,v}$} & 0.030 & 1.145\% & 0.141 & 5.252\% & 0.005 & 0.021\% \\
 \multicolumn{1}{l}{$p_{ij}$ (AE $\times 10^{-5}$)} & 1.204 & 0.025\% & 6.983 & 1.450\% & 0.043 & 0.009\% \\ 
 \midrule[0.3pt]\bottomrule[1pt]
 \end{tabular}
 \begin{tablenotes}
     \footnotesize
     \item Note: absolute error (AE) is $|a_i - b_i|,\forall i$ and absolute percentage error (APE) is $|a_i - b_i| / b_i ,\forall i$, where $a_i, b_i$ are estimations using our method and stochastic declustering, respectively. 
 \end{tablenotes}
 \end{threeparttable}
 }
 \end{table}

\end{document}